\documentclass{article}

\usepackage{amsmath,amsfonts,bm}

\def\eqref#1{equation~\ref{#1}}
\def\1{\bm{1}}

\def\rx{{\textnormal{x}}}
\def\ry{{\textnormal{y}}}

\def\vw{{\bm{w}}}

\DeclareMathAlphabet{\mathsfit}{\encodingdefault}{\sfdefault}{m}{sl}
\SetMathAlphabet{\mathsfit}{bold}{\encodingdefault}{\sfdefault}{bx}{n}

\def\sX{{\mathbb{X}}}
\def\sY{{\mathbb{Y}}}

\usepackage{hyperref}
\usepackage{url}

\usepackage{microtype}
\usepackage[pdftex]{graphicx}
\usepackage{booktabs} % for professional tables
\usepackage[font=small]{caption}
\usepackage[font=small]{subcaption}
\usepackage{bm}
\usepackage{enumitem}
\usepackage[ruled,vlined]{algorithm2e}
\usepackage{setspace}
\usepackage{titlesec}
\usepackage{physics}
\usepackage{chngpage}
\usepackage{dsfont}
\usepackage{etoolbox,lineno}

\usepackage{lipsum}
\usepackage{wrapfig}
\usepackage[accepted]{icml2024}

\usepackage{amsmath}
\usepackage{amssymb}
\usepackage{mathtools}
\usepackage{amsthm}
\usepackage{multirow}
\usepackage{multicol}

\usepackage[capitalize,noabbrev]{cleveref}

\theoremstyle{plain}
\newtheorem{theorem}{Theorem}

\newtheorem{lemma}[theorem]{Lemma}
\newtheorem{corollary}[theorem]{Corollary}

\newtheorem{remark}[theorem]{Remark}

\theoremstyle{definition}

\newcommand{\levi}{LEVI}

\definecolor{blush}{rgb}{0.87, 0.36, 0.51}
\definecolor{brickred}{rgb}{0.8, 0.25, 0.33}

\newcommand{\yuji}[1]{\textcolor{black}{#1}}

\icmltitlerunning{LEVI: Generalizable Fine-tuning via Layer-wise Ensemble of Different Views}

\begin{document}

\twocolumn[
% \icmltitle{LEVI: Layer-wise Ensemble of Different Views for Fine-tuning Generalization}
\icmltitle{LEVI: Generalizable Fine-tuning via Layer-wise Ensemble of Different Views}



\icmlsetsymbol{equal}{*}

\author{Yuji Roh\thanks{Work done during an internship at Google DeepMind.}$\,\,^{1}$, Qingyun Liu$^{2}$, Huan Gui$^{2}$, Zhe Yuan$^{3}$, Yujin Tang$^{2}$,\\
\textbf{Steven E. Whang}$^{1}$, \textbf{Liang Liu}$^{3}$, \textbf{Shuchao Bi}$^{3}$, \textbf{Lichan Hong}$^{2}$, \textbf{Ed H. Chi}$^{2}$, \textbf{Zhe Zhao}$^{2}$ \\
$^{1}$KAIST, \texttt{\{yuji.roh,swhang\}@kaist.ac.kr} \\
$^{2}$Google DeepMind, \texttt{\{qyl,hgui,yujintang,lichan,edchi,zhezhao\}@google.com}\\
$^{3}$Google Inc, \texttt{\{jeremyyuan,liangliu\}@google.com}\\
}

\begin{icmlauthorlist}
\icmlauthor{Yuji Roh}{google}
\icmlauthor{Qingyun Liu}{deepmind}
\icmlauthor{Huan Gui}{deepmind}
\icmlauthor{Zhe Yuan}{google}
\icmlauthor{Yujin Tang}{deepmind}
\\
\icmlauthor{Steven E. Whang}{kaist}
\icmlauthor{Liang Liu}{google}
\icmlauthor{Shuchao Bi}{google}
\icmlauthor{Lichan Hong}{deepmind}
\icmlauthor{Ed H. Chi}{deepmind}
\icmlauthor{Zhe Zhao}{deepmind}
\end{icmlauthorlist}

\icmlaffiliation{google}{Google Inc, Mountain View, CA, USA}
\icmlaffiliation{deepmind}{Google DeepMind, Mountain View, CA, USA}
\icmlaffiliation{kaist}{School of Electrical Engineering, Korea Advanced Institute of
Science and Technology (KAIST), Daejeon, Korea}

\icmlcorrespondingauthor{Yuji Roh}{yujiroh@google.com}

\icmlkeywords{Machine Learning, ICML}

\vskip 0.3in
]



\printAffiliationsAndNotice{}  % leave blank if no need to mention equal contribution

\begin{abstract}
Fine-tuning is becoming widely used for leveraging the power of pre-trained foundation models in new downstream tasks. 
While there are many successes of fine-tuning on various tasks, recent studies have observed challenges in the generalization of fine-tuned models to unseen distributions (i.e., out-of-distribution; OOD).
% While there are many success stories, generalizing the fine-tuned models to unseen distributions (i.e., out-of-distribution; OOD) remains a challenge according to recent studies. 
% Most of the existing solutions assume that a large pre-trained model is ``good enough'' and focus on fixing the issues in fine-tuning data only. However, we contend that the pre-trained model itself may also have problematic features that need to be suppressed. 
% We thus address both issues and propose \levi{}, a novel fine-tuning generalization method, where the pre-trained model is ensembled layer-wise with a smaller task-specific model, while preserving training and inference efficiencies. 
To improve OOD generalization, some previous studies identify the limitations of fine-tuning data and regulate fine-tuning to preserve the general representation learned from pre-training data. However, potential limitations in the pre-training data and models are often ignored. 
In this paper, we contend that overly relying on the pre-trained representation may hinder fine-tuning from learning essential representations for downstream tasks and thus hurt its OOD generalization. It can be especially catastrophic when new tasks are from different (sub)domains compared to pre-training data. 
To address the issues in both pre-training and fine-tuning data, we propose a novel generalizable fine-tuning method \levi{} \yuji{(\textbf{L}ayer-wise \textbf{E}nsemble of different \textbf{VI}ews)}, where the pre-trained model is adaptively ensembled layer-wise with a small task-specific model, while preserving its efficiencies. 
By combining two complementing models, \levi{} effectively suppresses problematic features in both the fine-tuning data and pre-trained model and preserves useful features for new tasks. Broad experiments with large language and vision models show that \levi{} greatly improves fine-tuning generalization via emphasizing different views from fine-tuning data and pre-trained features.
\vspace{-0.05cm}
% Through extensive experiments with large language and vision models, we show that \levi{} greatly improves the generalization of fine-tuned models via emphasizing the different views from fine-tuning data and pre-trained features.
\end{abstract}

% Previous version
% \begin{abstract}
% In the era of foundation models, fine-tuning has become a widely adopted strategy for leveraging the power of pre-trained models in new tasks. Despite many encouraging results of fine-tuning, recent studies have revealed that fine-tuned models still struggle to generalize in unseen distributions during training (i.e., out-of-distributions; OOD).
% Although a few algorithms have been proposed to improve generalization in fine-tuning, most existing solutions have primarily focused on the issues in the fine-tuning data without addressing the inherent limitations of the pre-trained models, which thus only partially solve the problem. 
% To address this issue, we propose \levi{}, which improves the fine-tuning generalization via an ensemble-based idea of jointly emphasizing the information from both large pre-trained and small task-specialized models, while preserving training and inference efficiencies. The key idea of \levi{} is to utilize the complementary views from two very different models to mitigate the impact of problematic features from not only fine-tuning data but also pre-trained features, while maintaining necessary features for the downstream task.
% As a result, \levi{} greatly improves the generalization ability of the fine-tuned models, as demonstrated in extensive experiments with large language models and vision models. 
% In this work, we show the importance of equally treating the information from both pre-trained features and fine-tuning data for improving fine-tuning generalization.
% \end{abstract}

\section{Introduction}
% \vspace{-0.05cm}

Recent breakthroughs in foundation models make various high-quality pre-trained models available to the public~\citep{devlin-etal-2019-bert, pmlr-v139-radford21a}, and the \textit{fine-tuning} paradigm has become a prevalent approach for leveraging pre-trained models' power in new downstream tasks. 
By tailoring the pre-trained features to align with the characteristics of the new tasks, fine-tuning shows promising performances in various scenarios, including natural language and computer vision~\citep{kornblith2019better, guu2020retrieval}.

\begin{figure*}[t]
% \vspace{-0.1cm}
\centering
\includegraphics[width=1\linewidth,trim=0cm 0.3cm 0cm 0cm]{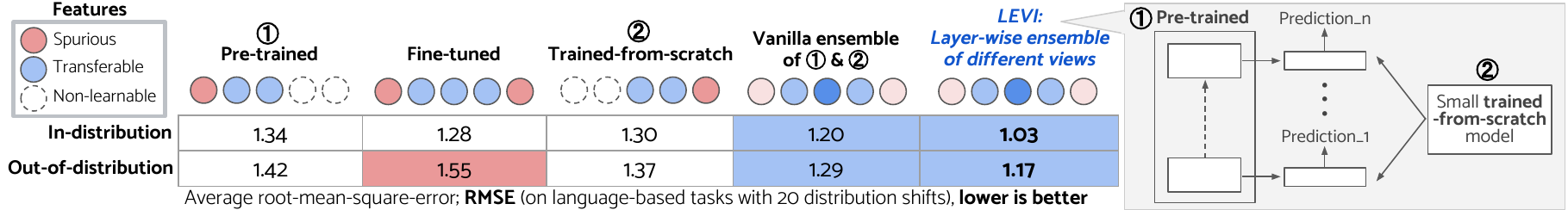}
\vspace{-0.45cm}
\caption{\small When both pre-trained features and fine-tuning data have inherent problems like spurious features, they can jointly affect the OOD generalization ability of a resulting fine-tuned model. Indeed, we observe that the OOD performance of the fine-tuned model is worse (\textcolor{brickred}{red} color in the table) than both the pre-trained and trained-from-scratch (i.e., randomly initialized then trained on fine-tuning data) models, where we 1) fine-tune a pre-trained language model (T5x) on various downstream tasks (movie and product recommendations) and 2) test on 20 distribution shifts (e.g., subpopulation and time shifts). 
To address this issue, our key idea is to \textit{separately leverage different views} from a pre-trained model and a trained-from-scratch model via \textit{layer-wise ensemble} to reduce the impact of problematic features while preserving necessary ones. Compared to the vanilla ensemble of such two complementing models (fourth column), \levi{} further improves both ID and OOD performances while preserving training and inference efficiencies -- see framework details in Sec.~\ref{sec:framework}.
% To address this issue, our key idea is to \textit{separately leverage different views} from a pre-trained model and a trained-from-scratch model via layer-wise ensemble to reduce the impact of problematic features while preserving necessary features, achieving great ID and OOD performances.
}
\vspace{-0.1cm}
\label{fig:overview}
\end{figure*}

Despite many success stories of fine-tuning, recent studies have observed that fine-tuned models often fail to ensure consistent generalization across new distributions (i.e., \textit{out-of-distribution}; OOD) at deployment time~\citep{Bommasani2021FoundationModels, kumar2022finetuning}, where OOD samples come from a different distribution than the data the model was fine-tuned on. 
% the one the model was trained on. 
Unlike traditional OOD generalization studies, fine-tuning pre-trained models faces unique challenges in improving OOD generalization, including computational costs of handling large models and the lack of access to pre-training data, which may already have inherent issues.
% where 1) the given pre-trained models may already have inherent issues, 2) the training and inference efficiency becomes more important as the model size increases, and 3) additional information on the pre-training or deployment (OOD) data is usually unavailable. 

% Concurrently, model generalization to unseen distributions (i.e., \textit{out-of-distribution}; OOD) remains a pivotal concern in machine learning research~\citep{tran2022plex}, where OOD samples come from a different distribution than the one the model was trained on. Given that pre-trained models are typically constructed on large and diverse data sources, there is an intuitive expectation for fine-tuned models based on such large pre-trained models to exhibit robust generalization capabilities when faced with OOD scenarios. Despite these anticipations, it is noteworthy that even fine-tuned models often fail to ensure consistent generalization across new distributions at deployment time~\citep{Bommasani2021FoundationModels, kumar2022finetuning}. 
% % especially when the downstream task is far from the pre-training distribution. 

Although several algorithms have been recently proposed to enhance fine-tuning generalization~\citep{kumar2022calibrated, kumar2022finetuning, wortsman2022robust, tian2023trainable}, most of them do not consider the inherent limitations in the pre-trained models. 
Specifically, many previous works implicitly assume that 1) the pre-trained models already have good enough features for the new tasks and that 2) any problems with generalization stem from the downstream (fine-tuning) data.
As a result, these algorithms mainly focus on preserving the original pre-trained features and avoiding overfitting to the fine-tuning data's problems, such as spurious features -- informative during training, but not useful (transferable) in general. 
However, such assumptions in prior works may not be true in practice, and pre-trained features can also have inherent issues.
For example, even if there is no spurious feature in the fine-tuning data, some of the features in the pre-trained model can be improperly used in the downstream tasks~\citep{xue2023eliminating}, e.g., pre-trained demographic features in language models can be wrongly used in new ranking systems. Moreover, pre-trained features may not have all the important representations for the new tasks~\citep{kang2023llms, Bommasani2021FoundationModels}; thus, simply preserving and relying on pre-trained representations may not be enough to learn essential ones for the new tasks.
% Recent works confirm that even large pre-trained models can be limited in adequately supporting new downstream tasks, as such models are still bounded by the data distributions used in pre-training~\citep{kang2023llms, Bommasani2021FoundationModels}.
% As an example, \citet{xue2023eliminating} show that even if there is no problematic feature in the fine-tuning data, some of the features in the pre-trained model can be improperly used in the target downstream task, which also results in a worse OOD generalization.

Therefore, we aim to \textit{mitigate the limitations from both pre-trained features and fine-tuning data while maintaining necessary features} to improve fine-tuning generalization. 
To systematically understand the problem, we use the spurious feature problem as a key example, as it is one of the most important factors causing OOD generalization failure. 
Notably, Fig.~\ref{fig:overview} shows that when both the pre-trained model and fine-tuning data contain spurious features,
the generalization ability of the fine-tuned model can be even worse than the pre-trained and trained-from-scratch (i.e., randomly initialized then trained on fine-tuning data) models, as the fine-tuned model can be affected by more spurious features than the other two. We provide other real-world cases in Sec.~\ref{sec:experiments}.
Here, the main challenge in the fine-tuning scenario is the lack of information about spurious features, especially from pre-training data that is unavailable during fine-tuning.

% Even though there are limitations in each of datasets, both pre-training and fine-tuning data can compensate each other's weaknesses. If we could leverage the pre-trained representations as well as the useful representations in fine-tuning datasets and use each other to compensate and eliminate the cause of spurious features.
% Even though there are limitations in each of them, we can leverage the useful representations from the two and mitigate the spurious features by comparing them.
Our key insight to tackle this challenge is using multiple views from both pre-training and fine-tuning data to compensate each other's weaknesses and keep useful features. 
We thus propose \levi{}, a \textbf{L}ayer-wise \textbf{E}nsemble for generalizable fine-tuning, which adaptively combines different \textbf{VI}ews from a large pre-trained model with those from a small, trained-from-scratch model to implicitly mitigate the impacts of spurious features and preserve necessary features, as described in the rightmost part of Fig.~\ref{fig:overview}. The fundamental insight of \levi{} is to harness the different views offered by such two complementary models. Compared to traditional fine-tuning that incrementally updates a pre-trained model on the fine-tuning data, we separate and jointly emphasize the complementary information from both sides. 
Also, as in Fig.~\ref{fig:overview}, \levi{} further improves performance and efficiency beyond simple prediction ensembles by tightly integrating the intermediate layers of the pre-trained model, which also offer different views (e.g., early layers are more general, and the later layers are more specific) -- see details in Sec.~\ref{sec:framework}.
% Also, to improve \levi{}'s performance and efficiency beyond simple prediction ensembling, we tightly integrate the intermediate layers of the pre-trained model, which also offer different views (e.g., early layers are more general, and the later layers are more specific), thereby enhancing the overall performance -- see details in Sec.~\ref{sec:framework}.

We note that there are several recent ensemble-based approaches for fine-tuning generalization~\citep{kumar2022calibrated, wortsman2022robust} that only utilize the pre-trained model variants, but we are the first to leverage the information from the trained-from-scratch models to mitigate the inherent problems from both pre-trained model and fine-tuning data and to generate new representations -- see detailed comparisons in Secs.~\ref{sec:relatedwork} \&~\ref{sec:experiments}. 
Especially for training from scratch, \levi{} uses a relatively small yet task-specialized model, which allows to decrease the computational costs of ensembling while giving additional benefits to effectively learn task-specialized features for new tasks.

Extensive experiments on language-based recommendation and computer vision tasks show that \levi{} greatly improves the generalization ability of fine-tuning. We observe the state-of-the-art results in various OOD scenarios, including subpopulation, time, and domain shifts. % while preserving training and inference efficiencies. 
% We observe the state-of-the-art performances in various OODs, including subpopulation, time, and domain shifts. 
We also show that our approach is more efficient than most existing ensemble-based generalization approaches. 
In addition, various ablation studies and hyperparameter analyses help to understand \levi{}'s behaviors. Finally, \levi{} can be gracefully merged with efficient fine-tuning methods such as LoRA~\citep{hu2022lora}, resulting in further improved training efficiency.

\vspace{-0.3cm}
\paragraph{Summary of Contributions:\hspace{-0.2cm}} 
(1) We reveal the importance of addressing inherent problems of not only fine-tuning data, but also pre-trained models via theoretical and empirical insights. (2) Based on such insights, we propose a novel layer-wise ensemble \levi{} for fine-tuning OOD generalization, which synergistically combines different views from two complementing models. (3) We show the value of leveraging trained-from-scratch models in mitigating the limitations of pre-trained models. (4) \levi{} largely improves OOD generalization in both language and vision models, while preserving training and inference efficiencies.

\vspace{-0.1cm}
\section{Related Work}
\label{sec:relatedwork}

\paragraph{OOD Generalization in the Traditional Literature\\} Among the broad model generalization issues, making the model more robust to various OOD scenarios becomes indispensable for AI deployment~\citep{shen2021towards}. 
% The main approaches for OOD generalization can be categorized into: 1) unsupervised representation learning that finds a better representation for diverse distributions~\citep{mahajan2021domain, zhang2022towards, 9088148},
% % ~\citep{mahajan2021domain, zhang2022towards, harary2022unsupervised, 9088148}, 
% 2) supervised model training that modifies model architectures or training processes to prevent the model from losing the generalization~\citep{finn2017model,li2018learning,d2019domain,carlucci2019domain, raghunathan2020understanding, xie2020n}, and 3) specialized optimization designing that uses robustness-aware objectives to ensure OOD performances~\citep{Sinha2017CertifyingSD,arjovsky2019invariant}.
Although there are many promising directions~\citep{finn2017model,li2018learning,d2019domain,carlucci2019domain, raghunathan2020understanding, roh2023improving}, many traditional studies do not consider the challenges in large models (e.g., inherent problems in the pre-trained features and scalability) or assume to access pre-training or unlabeled test data. In comparison, our work improves generalization in the fine-tuning pre-trained model paradigm, where 1) the given models may already have inherent issues, 2) the training and inference efficiency becomes more important, and 3) additional information on the pre-training or deployment (OOD) data is unavailable.
We leave a more detailed discussion on the traditional OOD literature in Sec.~\ref{appx:relatedwork}.

\vspace{-0.25cm}
\paragraph{OOD Generalization in the Fine-tuning Paradigm\\}
With the rapid growth of large pre-trained models, there is an emerging focus on OOD generalization in fine-tuning. 
% With the rapid growth of large pre-trained models, there is an emerging focus on improving OOD generalization in the fine-tuning pre-trained model setting. 

\vspace{-0.05cm}
As the pre-training data is known to cover large and diverse distributions, many recent approaches aim to preserve the pre-trained features for improving the generalization~\citep{kumar2022finetuning,tian2023trainable}. For example, \citet{kumar2022finetuning} propose a two-step approach that first linear probes the last layer and then fine-tunes all the parameters so as to mitigate the feature distortion during fine-tuning. Another work~\citep{tian2023trainable} designs a constrained bi-level optimization to minimally change the pre-trained features. In comparison, we consider the inherent problems in the pre-trained model itself, including the limited features for supporting downstream tasks and spurious correlations between pre-trained features and the target labels.
\yuji{Furthermore, to our understanding, we are the first to reveal the value of using trained-from-scratch models in mitigating inherent issues of pre-trained models.}

Among the fine-tuning generalization studies, recent ensemble approaches are the most relevant to our work as they utilize the information from multiple models for generalization and show promising improvements in computer vision tasks with various OOD scenarios~\citep{kumar2022calibrated, wortsman2022robust, pagliardini2023agree}. For example, \citet{kumar2022calibrated} average the outputs of the standard fine-tuned and robust-aware fine-tuned models. Another line of studies~\citep{wortsman2022robust,rame2022recycling,wortsman2022model} ensemble the model weights of the pre-trained model variants to gather diverse information -- e.g., ensembling fine-tuned and zero-shot models \yuji{or averaging multiple fine-tuned weights}.
Compared to these studies, \levi{} is the first work to emphasize the importance of separately treating the complementary views in pre-trained features and fine-tuning data, so as to also mitigate the inherent problems in the pre-trained features. % to enjoy both general and task-specialized views.
Another ensemble-based approach~\citep{pagliardini2023agree} trains multiple models to have disagreed predictions on the OOD distributions by accessing the unlabeled test data. In contrast, our work does not assume any information on the test OOD distribution. Moreover, \levi{} does not require more than one large model, preserving both training and inference efficiencies.

In addition, a recent work~\citep{xue2023eliminating} considers the spurious correlations in pre-trained models that can harm the group robustness and mitigates their impacts by utilizing a group-balanced dataset. However, this work assumes to have the information of spurious correlation in advance, which may not be available in real-world applications, especially when the pre-trained features and the target tasks become complex. In contrast, we do not use any prior know-ledge of spurious correlations in both the model and data.

\vspace{-0.1cm}
\section{When Fine-Tuning Fails to Generalize}
% \vspace{-0.2cm}

To improve fine-tuning generalization, we first explain how the inherent problems in the pre-trained model and fine-tuning data can jointly harm model generalization (Sec.~\ref{sec:theory}) and discuss the limitations of previous approaches (Sec.~\ref{sec:limitation}).
In this paper, we use the following settings and notations.

\vspace{-0.15cm}
\paragraph{Settings.}
We consider three data distributions each for pre-training, fine-tuning, and testing. 
Pre-training data and test data are not available, and we only have fine-tuning data and pre-trained model features. When we refer to training data, we mean the fine-tuning data.
We consider the fine-tuning distribution as \textit{in-distribution (ID)}, and the test distribution as \textit{out-of-distribution (OOD)}. Thus, ID data represents the samples that the model has been trained, and OOD data represents unfamiliar samples not seen during training.
% or unexpected

\vspace{-0.15cm}
\paragraph{Notations.}
Let $\rx \in \sX$, $\ry \in \sY$, and $\hat{\ry} \in \sY$ be the input feature, true label, and predicted label, respectively. Let $D$ be the data distribution.
Let $\vw$ be the model weights. The empirical loss is given by $L(\vw) {=} \frac{1}{m}\sum_{i} \ell(\ry_i,\hat{\ry}_i)$, where $m$ is the number of data samples, and $\ell(\cdot)$ is the loss function. 
% Let $D_\text{pretrain}$, $D_\text{id}$, and $D_\text{ood}$ be the pre-training, fine-tuning (ID), and test (OOD) distributions, respectively.

\subsection{Theoretical Backgrounds}
\label{sec:theory}

We first provide theoretical backgrounds that show the fine-tuned model can suffer from the spurious features from both pre-trained models and fine-tuning data.
Here, spurious features are defined as features that are useful to increase training (in-distribution) accuracy, but not transferable at deployment with new distributions. These features are known to hurt model generalization~\citep{pmlr-v119-sagawa20a, mccoy-etal-2019-right}. 
We note that the pre-trained models can have other inherent problems like demographic bias, but our work focuses on the issue directly affecting model generalization.

Traditionally, the influence of spurious features \textit{in training data} on model performance has received much attention. 
For example, \citet{nagarajan2021understanding} show that when there are statistical or geometric relationships between spurious features and labels, the empirical risk minimization (ERM)-based model can rely on such spurious correlations in prediction, resulting in a worse OOD generalization. Lemma~\ref{lemma:train_data_sc} summarizes the simple yet critical theoretical insights of previous work~\citep{nagarajan2021understanding, 10.1145/3442188.3445883}.
\begin{lemma}\label{lemma:train_data_sc}  ERM-based model training can be affected by spurious features in the training data. Let the training data $D$ has the input features $\rx = [x_1, x_2]$, where $x_1$ is a spurious feature and $x_2$ is a transferable feature. When we train a model with randomly initialized weights $\vw := [w_1, w_2]$ on $D$, the ERM-based trained model will have a non-zero $w_1$, indicating a spurious correlation was used to predict labels.
% , indicating that a spurious correlation was used to predict labels.
\end{lemma}

More recently, there is a new focus on investigating the potential harms from spurious correlations embedded \textit{within pre-trained features}. 
Specifically, \citet{xue2023eliminating} show that even if there is no spurious correlation in the fine-tuning data, the features in the pre-trained model can be spuriously used in the downstream task.
\begin{lemma}\label{lemma:pretrain_data_sc}(From~\citep{xue2023eliminating})
If a pre-trained model has spurious features for the downstream task, even the fine-tuning data that does not have spurious features may not eliminate the impacts of spurious correlations already embedded in the pre-trained model. Let the training data $D$ have the input features $\rx{=}[0, x_2]$, where $x_2$ is a transferable feature (i.e., no spurious feature). When we fine-tune a pre-trained model with non-zero weights $\vw := [w_1, w_2]$ on $D$, the resulting fine-tuned model will still have non-zero $w_1$.
% Even if there is no spurious correlation in the fine-tuning data, the features in the pre-trained model can be spuriously used in the downstream task.
\end{lemma}

Given the above lemmas, we can consider a case where a pre-trained model and fine-tuning data have different spurious features. Let the model weight $\vw$ be [$w_1$, $w_2$, $w_3$] and the data feature $\rx$ be [$x_1$, $x_2$, $x_3$], where $x_1$ and $x_3$ are spurious features for the downstream task. Here, a pre-trained model has an inherent spurious correlation with $x_1$ (i.e., nonzero $w_1$), and fine-tuning data contains only the spurious feature $x_3$, but not $x_1$ (i.e., [0, $x_2$, $x_3$]). When we train the given pre-trained model on the fine-tuning data, the model training will learn the spurious correlation from $x_3$ (Lemma~\ref{lemma:train_data_sc}), while not eliminating the pre-trained spurious correlation from $x_1$ (i.e., nonzero $w_1$), as the fine-tuning data is orthogonal to $x_1$ and cannot affect the model weight $w_1$ during training (Lemma~\ref{lemma:pretrain_data_sc}). As a result, we get the next corollary. 
\begin{corollary}\label{corollary:finetuned}
If both the pre-trained model and fine-tuning data have spurious features to the downstream labels, both spurious features will jointly affect the fine-tuned model. 
% Even if a pre-trained model and fine-tuning data do not contain spurious features of each other, the fine-tuned model will be jointly affected by both spurious features. 
% If both fine-tuning data and pre-trained models have spurious features to the labels, the fine-tuned model will be affected by both spurious correlations. 
\end{corollary}

\begin{remark}
Our work focuses on when spurious features in pre-training and fine-tuning data do not overlap. We note that when the two data share the same spurious features, the problem turns into a traditional spurious correlation study.
\end{remark}

\begin{figure}[h]
\vspace{-0.2cm}
\centering
\includegraphics[width=1.0\columnwidth,trim=0cm 0.3cm 0cm 0cm]{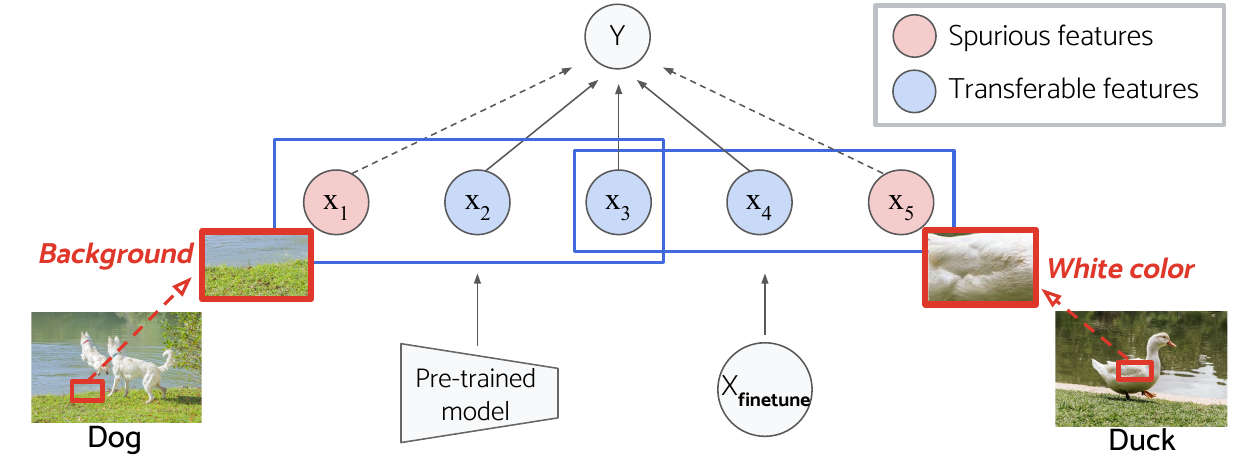}
\vspace{-0.4cm}
\caption{\small Toy example of a duck classification scenario.}
\vspace{-0.2cm}
\label{fig:toy_example}
\end{figure}

We provide a toy example in Fig.~\ref{fig:toy_example} to give more intuition on how both spurious features can jointly affect the fine-tuned model (Corollary~\ref{corollary:finetuned}), using a duck image classification scenario, where 1) the majority of duck images in fine-tuning data contain the white color of ducks, and 2) a given model is pre-trained to learn and focus on general vision features like backgrounds of the image.
Here, on the one hand, fine-tuning data may provide a white color bias during training, which can be used as spurious features derived from the data. As we discussed, many previous studies on fine-tuning generalization focus on such issues from fine-tuning data.
On the other hand, a pre-trained model itself may try to classify ducks using the background information based on its prior knowledge from pre-training data. 
Although the background information may be a good feature in general, it can become a spurious feature for this task, as ducks can be in various different places like ponds and grass, which are also accessible to other animals. Hence, the feature may not generalize well to test data containing such OODs.

% We provide a toy example in Fig.~\ref{fig:toy_example} to give more intuition on how both spurious features can jointly affect the fine-tuned model (Corollary~\ref{corollary:finetuned}), using a duck image classification scenario, where 1) the majority of duck images in fine-tuning data contain water backgrounds, and 2) a given model is pre-trained to learn and focus on general vision features like object color.
% Here, on the one hand, fine-tuning data may provide a water background bias during training, which can be used as spurious features derived from the data. As we discussed, many previous studies on fine-tuning generalization focus on such issues from fine-tuning data.
% On the other hand, a pre-trained model itself may try to classify ducks using the object's color information based on its prior knowledge from pre-training data. 
% Although the object color may be a good feature in general, it can become a spurious feature for this task as there are non-white ducks and other animals that are white (e.g., dolphins). Hence, the feature may not generalize well to test data containing such OODs.

% \begin{wrapfigure}{r}{7.9cm}
% \vspace{-0.7cm}
% \centering
% \includegraphics[width=0.5\columnwidth,trim=0cm 0.3cm 0cm 0cm]{figures/toy_example.pdf}
% \caption{\small Toy example.}
% \vspace{-1.cm}
% \label{fig:toy_example}
% \end{wrapfigure}
Thus, as the fine-tuned model can be affected by more spurious correlations (e.g., both $x_1$ and $x_5$ in Fig.~\ref{fig:toy_example}) than the pre-trained model and the trained-from-scratch model (i.e., randomly initialized then trained on fine-tuning data), the generalization ability of the fine-tuned model can be even worse than the other two. This insight also aligns well with Fig.~\ref{fig:overview} -- see a more concrete synthetic example in Sec.~\ref{appx:synthetic_ex} and empirical results on real-world benchmarks in Sec.~\ref{sec:experiments}.

\vspace{-0.1cm}
\subsection{Limitation of Previous Approaches}
\label{sec:limitation}

% As we discussed above, mitigating the influence of spurious features on both fronts is essential for enhancing the generalization ability of the fine-tuning pre-trained model paradigm. 
Despite the importance of mitigating the influence of inherent issues on both fronts for enhancing fine-tuning generalization, it has not been actively considered in previous studies.
% it is barely considered in previous studies. 
Most existing studies for fine-tuning generalization challenge the long-held assumptions that pre-trained model features are both 1) free from inherent flaws and 2) good enough to support new tasks. Their strategies thus focus on preserving the original pre-trained features during fine-tuning and avoiding overfitting to the fine-tuning data. 

However, these assumptions may not hold in reality. First, pre-trained features may contain inherent problems~\citep{Bommasani2021FoundationModels, xue2023eliminating}, which can also limit OOD generalization, as observed above. Also, pre-trained features may not contain all necessary information for new tasks, thus simply preserving the pre-trained features cannot be the best solution to adequately support such new tasks.

\vspace{-0.1cm}
\section{Framework}
\label{sec:framework}

We now design a new fine-tuning framework for improving generalization in OODs. 
We follow the previous theoretical insights, which show the importance of reducing the impact of spurious features from both pre-trained features and fine-tuning data while maintaining necessary features.

\vspace{-0.1cm}
\paragraph{Key Intuition.}
Our main idea is to leverage \textit{different views from different models} to mitigate problems in both pre-trained features and fine-tuning data, while learning essential representations.
Despite the importance of addressing spurious correlations during fine-tuning, we usually do not have any information about them, especially those from the pre-training distribution. It is thus hard to explicitly prevent utilizing spurious correlations during fine-tuning.
Instead, we implicitly mitigate such spurious correlations by merging complementary information from two very different models: a pre-trained model and a trained-from-scratch model. These models will not be affected by the spurious features from each other, thus we expect to reduce the impact of such features by comparing the complementing information between them. Here, a key difference from conventional fine-tuning is that we separate and jointly emphasize the signals from pre-training and fine-tuning (downstream) data.
%, while fine-tuning incrementally updates a pre-trained model on the data.

\vspace{-0.15cm}
\paragraph{Using Ensemble.}
% Model ensembling is an effective way to combine information from diverse, independent models. Ensembling reduces model variance and improves generalization error. For our research, we will ensemble distinct models as a starting point to integrate their complementary strengths.
A key question is how we can effectively combine the distinct information from different models. Our starting point is the idea of model ensemble, which is known to be very effective in combining information from multiple models~\citep{zhang2012ensemble}. Especially, when the models are diverse and independent, ensembling is a good choice to reduce the model variance, which can decrease the generalization error~\citep{kotu2014predictive}. 
% Therefore, given our goal to utilize two distinctly different models, adopting the concept of an ensemble approach can be a reasonable starting point.
% (i.e., making the final model less sensitive to new inputs)

% Using multiple large models can face a critical concern that the training and inference costs significantly increase.
% A common way to perform an ensemble of different models can be a simple (weighted) average between their final predictions~\citep{GANAIE2022105151}. 

% Indeed, when we perform the final prediction ensemble with the pre-trained and trained-from-scratch models to follow our intuition, the generalization ability is clearly improved in various OOD cases -- see empirical results in Sec.~\ref{sec:experiments}. 
% However, such a simple ensemble faces a critical concern that the training and inference costs significantly increase, especially when the two models are both large.

\vspace{-0.15cm}
\paragraph{Methodology.}
Our design focuses on two key objectives: 1) harnessing a multitude of diverse views to maximize the benefits of ensembling beyond simply averaging final model predictions, and 2) maintaining efficiencies in both training and inference phases, which is often a critical issue in ensemble methods. 
We achieve these objectives by leveraging different views from the models in two directions: \textit{among} the models and \textit{within} the model. First, by ensembling pre-trained and trained-from-scratch models, we can obtain a general view from pre-training data and a task-specialized view from fine-tuning data. Also, by ensembling within the model, we can utilize different information from each intermediate layer (e.g., early layers are more general, and the last layers are more specific)~\citep{yosinski2014transferable}. Note that using multiple intermediate layers is also known to be beneficial in fine-tuning generalization~\citep{evci2022head2toe}.

\begin{figure}[t]
% \vspace{-0.1cm}
\centering
\includegraphics[width=0.95\columnwidth,trim=0cm 0.3cm 0cm 0cm]{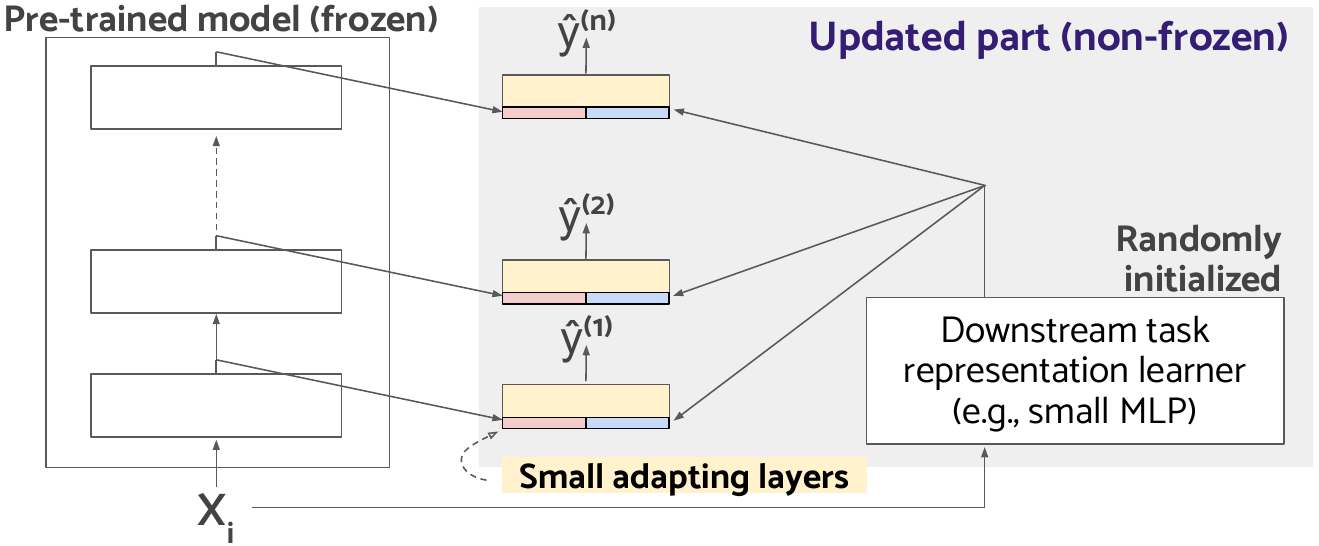}
\vspace{-0.3cm}
\caption{\small \levi{} architecture of using a layer-wise ensemble.}
% \caption{\levi{} architecture of using a wide-and-deep ensemble.}
\label{fig:genie}
\vspace{-0.45cm}
\end{figure}

By this intuition, we propose \levi{}, a layer-wise ensemble framework for fine-tuning generalization. 
As described in Fig.~\ref{fig:genie}, \levi{} tightly integrates 1) a pre-trained model, which does not need to be updated during fine-tuning, together with 2) a small randomly-initialized model by utilizing adapting layers that concatenate two models.
% \levi{} tightly integrates a pre-trained model together with a small randomly-initialized model by utilizing adapting layers, as shown in Fig.~\ref{fig:genie}. \levi{} uses two models:
% 1) a pre-trained model, which does not need to be updated during fine-tuning, and 2) a relatively small, randomly initialized model.
Here, the role of the randomly initialized model is to learn necessary and specialized representations from the fine-tuning data. 
Also, the model size can be much smaller than the pre-trained model.
We then connect the two models with small adapting layers, where their inputs concatenate 1) the pre-trained model’s intermediate outputs and 2) the randomly-initialized model final output for each input data.
As a result, we can tightly ensemble the information from both pre-trained and trained-from-scratch (i.e., randomly initialized then trained) models.
Finally, we set the adapting layer outputs to be the predicted labels and update the model via the following loss function: 
\vspace{-0.1cm}
\setlength{\abovedisplayskip}{0pt}
\setlength{\belowdisplayskip}{2pt}
\begin{gather*}
\min_{\vw} \frac{1}{m} \cdot \frac{1}{n} ~ \overset{m}{\underset{i}{\text{\small$\sum$}}} ~ \overset{n}{\underset{j}{\text{\small$\sum$}}} ~ \ell(\ry_i,\hat{\ry}^{(j)}_i), 
\end{gather*}
where $m$ and $n$ are the numbers of training samples and adapting layers, respectively, and $\hat{\ry}^{(j)}_i$ is the predicted label from the adapting layer $j$ for the input data $i$. 
Note that we consider the equal weights on all $\hat{\ry}^{(j)}$, and one can change it as a weighted sum of $\hat{\ry}^{(j)}$ -- see more discussions in Sec.~\ref{appx:weight_sum}.

\vspace{-0.1cm}
\paragraph{Benefits.}
Finally, we highlight two key benefits of \levi{}.

\vspace{-0.05cm}
\textit{Adding a Task-specialized Model:} 
\levi{} has the flexibility to use any model architecture as the trained-from-scratch model, which allows for the selection of task-specialized architectures. This flexibility is very beneficial to learn critical features for new tasks effectively, which is particularly notable when the pre-trained features are not enough to support the new tasks. 
We provide \textit{a concrete example in movie recommendation}, aiming to predict user ratings for movies using a pre-trained language model. \levi{} can utilize a multi-layer perceptron (MLP) model with an embedding block as its trained-from-scratch model. Here, such an MLP model is known to be very suitable for handling sparse user and movie information, offering a distinct advantage over conventional language models, thereby enhancing the overall recommendation performances.

\textit{Using Multiple Intermediate Layers:}
Another advantage of \levi{} comes from leveraging both the later and early layers of a large model, which contributes to enhancing the trade-off between in-distribution (ID) and out-of-distribution (OOD) performances.
In our experiments, we observe notable results indicating that the later layers of the large model enhance ID performance, while the early layers contribute positively to OOD performance -- see detailed results in Sec.~\ref{sec:algorithm_analysis}. This intuition aligns with the previous knowledge in the literature that different levels of intermediate layers in a model have different characteristics, e.g., early layers are more general, and later layers are more specific~\citep{yosinski2014transferable}.
\levi{} uses this knowledge to improve overall results, as it enjoys different levels of intermediate outputs.

\vspace{0.05cm}
\begin{remark}[Using a Fine-tuned Large Model] \label{remark:using_adapted}
We note that \levi{}'s pre-trained model itself can also be replaced with an adapted (i.e., fine-tuned) model.
% Note that \levi{} is flexible to replace the pre-trained model with an adapted (i.e., fine-tuned) model. 
When the given large model is suitable to learn the target downstream task (e.g., using ImageNet pre-trained ViT for ImageNet-related tasks), using the adapted model can be helpful to further improve overall performances of \levi{}.
% We note that \levi{} is flexible to replace the pre-trained model with an adapted (i.e., fine-tuned) model to further improve the overall performances, especially when the given large model is very suitable to learn the target downstream task. 
% While the primary setting of \levi{} is to use an original pre-trained model, we can replace it with an adapted (i.e., fine-tuned) model to further improve the overall performances, especially when the given large model is very suitable to learn the target downstream task. 
In such cases, \levi{}'s training efficiency is still better than ensembling multiple large models and comparable with fine-tuning one large model.
\end{remark}

% \vspace{0.2cm}
\begin{remark}[Compatibility with Efficient Training Approaches]
% Recently, several efficient training methods for large models have been proposed to reduce computational costs while maintaining reasonable performances~\citep{hu2022lora, pmlr-v119-chen20s}. For example, LoRA~\citep{hu2022lora} utilizes rank decomposition matrices in Transformer layers to reduce the number of training parameters. 
If we replace \levi{}'s pre-trained model with an adapted (i.e., fine-tuned) model, we can also utilize
efficient training methods for large models~\citep{hu2022lora, pmlr-v119-chen20s} like LoRA, which further enhance the efficiency of \levi{} -- see results in Sec.~\ref{sec:compatibility}.
\end{remark}

\vspace{-0.1cm}
\section{Experiments}
\label{sec:experiments}
% \vspace{-0.05cm}

We perform extensive experiments to evaluate \levi{} in various scenarios. All experiments are repeated with three random seeds. We use TensorFlow~\citep{tensorflow2015-whitepaper} with JAX~\citep{jax2018github} and Flax~\citep{flax2020github}. More detailed settings (e.g., hyperparameters) are in Sec.~\ref{appx:settings}.

\vspace{-0.1cm}
\paragraph{Datasets \& OOD Scenarios.} We consider various OOD scenarios in two modalities: language-based recommendation tasks and computer vision tasks. For language-based recommendation, we use MovieLens~\citep{10.1145/2827872} and Amazon Review~\citep{ni2019justifying} \yuji{-- see an additional experiment for another natural language processing task in Sec.~\ref{appx:nlp}.}
For vision, we use Diabetic Retinopathy (Medical)~\citep{diabetic-retinopathy-detection} and ImageNet-Variants~\citep{wang2019learning, hendrycks2021many, hendrycks2021nae, recht2019imagenet}. All datasets are from TensorFlow Datasets~\citep{TFDS}. Pre-processing details are in Sec.~\ref{appx:preprocessing}.

\vspace{-0.05cm}
\textit{[MovieLens]} Contains movie rating data from online movie website, where each data sample has 12 attributes (e.g., user id, movie title, and movie genre). 
We utilize user-id, user-age, user-occupation, user-zipcode, movie-title, and movie-id as input features and rating as the label attribute, where the rating range is $[1, 5]$. 
% We utilize \texttt{user-id}, \texttt{user-age}, \texttt{user-occupation}, \texttt{user-zipcode}, \texttt{movie-title}, and \texttt{movie-id} as input features and \texttt{rating} as the label attribute, where the rating range is $[1, 5]$. 
% We use a stable version of MovieLens that contains 100,000 ratings from 943 users on 1,682 movies. 
% Each user gives ratings on at least 20 movies. 
% For the OOD scenario, we consider genre shifts, where we train all algorithms on data with top-5 popular genres (action, comedy, drama, romance, thriller) and test on data with other genres (e.g., animation, SciFi).
For the OOD scenario, we consider \textbf{genre (subpopulation) shifts}: the ID data contains movies of top-5 popular genres (action, comedy, drama, romance, thriller), and the OOD data contains other 12 genres (e.g., animation, sci-fi) that have at least 200 data points.

\vspace{-0.05cm}
\textit{[Amazon Review]} Contains product rating data from the Amazon.com website, where each data sample has 15 attributes (e.g., customer id, product category, product title). We utilize customer-id, product-title, and product-id as input features and rating as the label attribute, where the rating range is $[1, 5]$.
For the OOD scenarios, we consider \textbf{time shifts} and \textbf{product (subpopulation) shifts}: the ID data contains the first 4 years' (oldest) ratings of books, and the OOD data contains the most recent year's ratings of books and other products (e.g., watch, toy, sports, jewelry).

\vspace{-0.05cm}
\textit{[Diabetic Retinopathy (Medical)]} Contains human retina images, where the label attribute is the severity of diabetic retinopathy in the range $[0, 4]$. 
For the OOD scenario, we consider \textbf{quality shifts}: the ID data and OOD data contain high-resolution and low-resolution images, respectively.

\vspace{-0.05cm}
\textit{[ImageNet-Variants]} Contains different styles (e.g., sketch, adversarial) of ImageNet datasets with 1,000 label classes. For the OOD scenario, we consider \textbf{domain shifts}: the ID dataset is ImageNet-Sketch, and the OOD datasets are ImageNet-A, ImageNet-R, and ImageNet-V2.

\begin{table*}[t]
  \caption{Performances on the MovieLens and Amazon Review datasets. All algorithms are evaluated on separate ID and OOD datasets using the root-mean-square error (RMSE ${=} \sqrt{\sum_i{(\ry_i{-}\hat{\ry}_i)^2}/m}$), a standard metric for recommendation systems, where lower is better. % Note that RMSE is a standard metric for recommendation systems.
% \vspace{-0.2cm}
  }
  \label{tbl:recom_main}
  \centering
\scalebox{0.82}{
% \hspace{-0.3cm}
  \begin{tabular}{cl@{\hspace{12pt}}c@{\hspace{9pt}}c@{\hspace{12pt}}c@{\hspace{9pt}}c@{\hspace{9pt}}c}
    \toprule
      & & \multicolumn{2}{c}{MovieLens} & \multicolumn{3}{c}{Amazon Review} \\
    \cmidrule(lr){1-2}\cmidrule(lr){3-4}\cmidrule(lr){5-7}
      & Method &  ID & OOD (12 genres) & ID & OOD (time) & OOD (7 products) \\
    \cmidrule(lr){1-2}\cmidrule(lr){3-4}\cmidrule(lr){5-7}
    \multirow{4}{*}{\shortstack{Standard\\Training\\Baselines}} & T5x Fine-tuning (FT) & 1.175\tiny{$\pm$0.021} & 1.268\tiny{$\pm$0.036} & 1.386\tiny{$\pm$0.024} & 1.543\tiny{$\pm$0.014} & 2.029\tiny{$\pm$0.154} \\
    % \cmidrule(lr){1-1}\cmidrule(lr){2-3}\cmidrule(lr){4-5}
    & T5x Light-tuning: Half of transformers (HT) & 1.230\tiny{$\pm$0.006} & 1.297\tiny{$\pm$0.005} & 1.436\tiny{$\pm$0.035} & 1.529\tiny{$\pm$0.014} & 1.862\tiny{$\pm$0.289} \\
    & T5x Light-tuning: Linear probing (LP) & 1.226\tiny{$\pm$0.014} & 1.218\tiny{$\pm$0.054} & 1.453\tiny{$\pm$0.065} & 1.529\tiny{$\pm$0.045} & 1.759\tiny{$\pm$0.130} \\
    & T5x From scratch (FS) & 1.175\tiny{$\pm$0.021} & 1.291\tiny{$\pm$0.019} & 1.410\tiny{$\pm$0.064} & 1.532\tiny{$\pm$0.014} & 1.494\tiny{$\pm$0.010}\\
    \cmidrule(lr){1-2}\cmidrule(lr){3-4}\cmidrule(lr){5-7}
    \multirow{4}{*}{\shortstack{Fine-tuning\\Generalization\\Baselines}} & LP$\rightarrow$FT~\citep{kumar2022finetuning} & 1.163\tiny{$\pm$0.014} & 1.357\tiny{$\pm$0.009} & 1.407\tiny{$\pm$0.003} & 1.503\tiny{$\pm$0.015} & 1.669\tiny{$\pm$0.025} \\
    & FT+RobustModel~\citep{kumar2022calibrated} & 1.072\tiny{$\pm$0.002} & 1.127\tiny{$\pm$0.038} & 1.374\tiny{$\pm$0.016} & 1.498\tiny{$\pm$0.004} & 1.752\tiny{$\pm$0.130} \\
    & FT+FS & 1.058\tiny{$\pm$0.011} & 1.159\tiny{$\pm$0.018} & 1.333\tiny{$\pm$0.019} & 1.504\tiny{$\pm$0.006} & 1.521\tiny{$\pm$0.029} \\
    & FT+ZeroShot~\citep{wortsman2022robust} & 1.177\tiny{$\pm$0.010} & 1.253\tiny{$\pm$0.038} & 1.336\tiny{$\pm$0.015} & 1.517\tiny{$\pm$0.017} & 2.515\tiny{$\pm$0.175} \\
    \cmidrule(lr){1-2}\cmidrule(lr){3-4}\cmidrule(lr){5-7}
    & \levi{} & \textbf{0.932\tiny{$\pm$0.005}} & \textbf{1.065\tiny{$\pm$0.018}} & \textbf{1.095\tiny{$\pm$0.003}} & \textbf{1.310\tiny{$\pm$0.006}} & \textbf{1.298\tiny{$\pm$0.006}} \\
    % \levi{} with original T5x & 0.944\tiny{$\pm$0.004} & 1.104\tiny{$\pm$0.003} & 1.109\tiny{$\pm$0.003} & \textbf{1.296\tiny{$\pm$0.011}}\\
    % \levi{} with light-tuned T5x & 0.932\tiny{$\pm$0.005} & \textbf{1.065\tiny{$\pm$0.018}} & 1.095\tiny{$\pm$0.003} & 1.310\tiny{$\pm$0.006}\\
    % \levi{} with full fine-tuned T5x & \textbf{0.927\tiny{$\pm$0.004}} & 1.085\tiny{$\pm$0.014} & \textbf{1.093\tiny{$\pm$0.001}} & 1.318\tiny{$\pm$0.004}\\
    \bottomrule
  \end{tabular}
  }
  \vspace{-0.25cm}
\end{table*}

% \vspace{-0.15cm}
\paragraph{Models.}
We use two large pre-trained models: T5x~\citep{2020t5, roberts2022t5x} and ImageNet-21k pre-trained ViT~\citep{dosovitskiy2020vit} for language-based recommendation and computer vision tasks, respectively.
\yuji{We note that LEVI can also gracefully work together with other types of model backbones, as LEVI is a model-agnostic approach for improving OOD generalization.}

In \levi{}, we use a small randomly-initialized model and adapting layers together with the pre-trained model, as explained in Fig.~\ref{fig:genie}. For the small model, we use a two-layer multi-layer perceptron (MLP) model with input embedding layers for recommendation tasks and a four-layer convolutional neural network (CNN) model for computer vision tasks. 
% We note that the small randomly-initialized model in \levi{} can be defined to effectively learn the task-specialized features, and we follow the general knowledge from recommendation and computer vision literature.
Each adapting layer is composed of an MLP with one hidden layer. Details on the model configurations (e.g., number of neurons in hidden layers) can be found in Sec.~\ref{appx:model_settings}.

% \vspace{-0.15cm}
\paragraph{Baselines.} 
We compare \levi{} with three types of baselines: 1) standard training baselines for pre-trained models, 2) state-of-the-art fine-tuning generalization baselines, and 3) parameter-efficient fine-tuning baselines.

For standard training, we consider four baselines: full fine-tuning (FT), light-tuning of half of the parameters (i.e., half of transformers; HT), light-tuning of the last linear layer (i.e., linear probing; LP), and training-from-scratch (FS). The full fine-tuning baseline updates all pre-trained model parameters, while the light-tuning baselines update them partially. These light-tuning baselines are also known as robust fine-tuning methods for OODs~\citep{kumar2022calibrated}. The training-from-scratch baseline first randomly initializes all parameters then trains on the fine-tuning data.

For fine-tuning generalization, we consider four state-of-the-art baselines for large pre-trained models: LP$\rightarrow$FT~\citep{kumar2022finetuning}, FT+RobustModel~\citep{kumar2022calibrated}, FT+ZeroShot~\citep{wortsman2022robust}, and FT+FS. 
LP$\rightarrow$FT is a two-step baseline that first performs linear probing then full fine-tuning to mitigate the pre-trained feature distortion originated from the randomly initialized head. FT+RobustModel, FT+ZeroShot, and FT+FS are the ensemble baselines. FT+RobustModel ensembles the calibrated outputs of a fine-tuned model and a robustness-aware trained model, which are assumed to achieve good ID and OOD performances, respectively. Here we use linear probing (LP) as the robustness-aware model, as in the original paper~\citep{kumar2022finetuning}. FT+ZeroShot ensembles the parameters of a fine-tuned model and a pre-trained (zero-shot) model to preserve general features in the pre-trained model. FT+FS ensembles the outputs of a fine-tuned model and a trained-from-scratch model to reduce the impacts of problematic pre-trained features via the trained-from-scratch model.

For parameter efficient fine-tuning, we utilize LoRA~\citep{hu2022lora}, a state-of-the-art method that replaces Transformer layers with rank decomposition matrices to reduce the number of training parameters. 
We use LoRA in Sec.~\ref{sec:compatibility} to further improve \levi{}'s training efficiency.
% This baseline is used in Sec.~\ref{sec:compatibility} to further improve the training efficiency of \levi{} via LoRA.

% \paragraph{Metrics}
% We focus on two accuracy metrics and two efficiency metrics. \textit{[Accuracy]} To evaluate the performances of algorithms on ID and OOD data, we use 1) root-mean-square error (RMSE ${=} \sqrt{\sum_i{(\ry_i{-}\hat{\ry}_i)^2}/m}$) for recommendation tasks and 2) standard accuracy over all samples for computer vision tasks. Note that RMSE is a standard metric for evaluating recommendation systems. \textit{[Efficiency]} We compare 1) number of model parameters in training and inference and 2) floating point operations (FLOPs), which are widely used to estimate required memories and computational costs.

% \vspace{-0.1cm}
\subsection{Performances on IDs and OODs}
\label{sec:performances_id_ood}
% \subsection{Performances on In-Distributions and Out-of-Distributions}

% \vspace{-0.05cm}
\paragraph{Recommendation.}

Table~\ref{tbl:recom_main} shows the in-distribution (ID) and out-of-distribution (OOD) performances on the MovieLens and Amazon Review datasets. Here we consider the following OOD scenarios: genre, time, and product shifts. The full results for genre and product shifts are in Sec.~\ref{appx:recom_all}.
% in Tables~\ref{tbl:movielens_genre} and~\ref{tbl:amazon_product}.

For the standard training baselines, full fine-tuned models indeed show worse OOD performances than at least one of all the light-tuned (e.g., linear-probed) and trained-from-scratch models, as expected in Sec.~\ref{sec:theory}. Notably, for product shifts in Amazon Review (last column of Table~\ref{tbl:recom_main}), the fine-tuned model performs far worse than all the light-tuned and trained-from-scratch models -- full results are in Table~\ref{tbl:amazon_product}.

The state-of-the-art fine-tuning generalization baselines mostly improve the OOD performances of the fine-tuned model. Among these baselines, we observe that the heavy ensemble approaches FT+RobustModel and FT+FS show promising performances. Also, we note that the FT+ZeroShot baseline, which has a core assumption that the pre-trained (zero-shot) features can give generally good features to the target task, sometimes largely fails to improve the OOD performances in our recommendation tasks (e.g., last column of Table~\ref{tbl:recom_main}). We suspect that the pre-trained (zero-shot) language features are not sufficient for this downstream task and thus may even harm the results.

In contrast, \levi{} greatly improves both ID and OOD results in all scenarios by combining two complementing models. This result shows that \levi{} effectively suppresses the problematic features in both the fine-tuning data and pre-trained model, while preserving useful features for the task. 

% \yuji{We also perform an experiment for sentiment classification, a different type of NLP task, and observed consistent performance improvements when using LEVI.}

\begin{table}[t]
\vspace{-0.2cm}
  \caption{Performances on the Medical and ImageNet datasets using standard accuracy, where higher is better. We mark the best and second best results with \textbf{bold} and \underline{underline}, respectively.
%   All algorithms are evaluated on separate ID and OOD datasets using standard accuracy over all samples, where higher is better.
% \vspace{-0.2cm}
  }
  \label{tbl:vision_main}
  \centering
\scalebox{0.77}{
% \hspace{-0.3cm}
  \begin{tabular}{l@{\hspace{15pt}}c@{\hspace{9pt}}c@{\hspace{12pt}}c@{\hspace{9pt}}c}
    \toprule
      & \multicolumn{2}{c}{Medical} & \multicolumn{2}{c}{ImageNet} \\
    \cmidrule(lr){1-1}\cmidrule(lr){2-3}\cmidrule(lr){4-5}
      Method &  ID Acc & OOD Acc & ID Acc & OOD Acc \\
    \cmidrule(lr){1-1}\cmidrule(lr){2-3}\cmidrule(lr){4-5}
    ViT Fine-tuning (FT) & 78.58\tiny{$\pm$0.16} & 74.20\tiny{$\pm$0.15} &
    83.06\tiny{$\pm$0.05} & 21.62\tiny{$\pm$0.48}\\
    % \cmidrule(lr){1-1}\cmidrule(lr){2-3}\cmidrule(lr){4-5}
    ViT Light-tuning: HT & 77.04\tiny{$\pm$0.31} & 73.63\tiny{$\pm$0.04} & 81.52\tiny{$\pm$0.17} & 28.85\tiny{$\pm$0.30}\\
    ViT Light-tuning: LP & 74.76\tiny{$\pm$0.08} & 71.21\tiny{$\pm$0.38} & 66.48\tiny{$\pm$0.11} & 31.49\tiny{$\pm$0.07}\\
    \cmidrule(lr){1-1}\cmidrule(lr){2-3}\cmidrule(lr){4-5}
    LP$\rightarrow$FT & \textbf{78.83\tiny{$\pm$0.02}} & 73.81\tiny{$\pm$0.03} & 81.39\tiny{$\pm$0.14} & 28.26\tiny{$\pm$0.16}\\
    FT+RobustModel & 78.16\tiny{$\pm$0.01} & 74.16\tiny{$\pm$0.07} & \textbf{84.22\tiny{$\pm$0.15}} & 30.30\tiny{$\pm$0.26}\\
    FT+ZeroShot & 78.35\tiny{$\pm$0.46} & \underline{74.36\tiny{$\pm$0.10}} &
    82.70\tiny{$\pm$0.13} &
    \underline{32.10\tiny{$\pm$0.16}}\\
    \cmidrule(lr){1-1}\cmidrule(lr){2-3}\cmidrule(lr){4-5}
    \levi{} & \underline{78.60\tiny{$\pm$0.04}} & \textbf{75.15\tiny{$\pm$0.05}} & \underline{83.33\tiny{$\pm$0.03}} & \textbf{33.62\tiny{$\pm$0.12}}\\
    % \levi{} with original ViT & 76.45\tiny{$\pm$0.22} & 73.63\tiny{$\pm$0.02} & xx.xx\tiny{$\pm$xx.xx} & xx.xx\tiny{$\pm$xx.xx}\\
    % \levi{} with light-tuned ViT & 77.03\tiny{$\pm$0.17} & 73.67\tiny{$\pm$0.06} &
    % 81.62\tiny{$\pm$0.34} & 31.40\tiny{$\pm$0.07}\\
    % \levi{} with full fine-tuned ViT & \underline{78.60\tiny{$\pm$0.04}} & \textbf{75.15\tiny{$\pm$0.05}} & \underline{83.33\tiny{$\pm$0.03}} & \textbf{33.62\tiny{$\pm$0.12}}\\
    \bottomrule
  \end{tabular}
  }
  \vspace{-0.35cm}
\end{table}

% \vspace{-0.1cm}
\paragraph{Image Classification.}

Table~\ref{tbl:vision_main} shows the ID and OOD performances on the Medical and ImageNet-variants datasets with quality shifts and domain shifts, respectively. The full results for ImageNet OOD shifts are in Table~\ref{tbl:imagenet}. We note that as ViT models are known to be hard to train on small or mid-sized training data using random weight initialization~\citep{steiner2022how}, we do not use the training-from-scratch baseline in ViT experiments.
Interestingly, in the medical dataset, the robust training baselines including light-tuning and LP$\rightarrow$FT are not helpful in improving generalization. The major assumption of these approaches is that the pre-trained features have ``good-enough'' features for supporting downstream tasks, but the ImageNet pre-trained ViT may not have enough features to support the medical image-specific features.
On the other hand, in the ImageNet classification, as the ImageNet pre-trained ViT can be considered to already have reasonable features to support ImageNet-variants OOD datasets, the previous robust training baselines indeed improve the generalization of the fine-tuned model.
In both datasets, \levi{} can achieve the best OOD accuracies among all baselines while having ID accuracies comparable to that of the fine-tuned model.

% \vspace{-0.1cm}
\subsection{Efficiency Comparison}
\label{sec:efficiency}

We compare the efficiency of algorithms in Table~\ref{tbl:efficiency} varying 1) the number of model parameters and 2) floating point operations (FLOPs), which are widely used to estimate required memories and computational costs. When evaluating \levi{}, we either use a pre-trained model or its fine-tuned version. 
% \levi{} has a flexibility of using either original pre-trained model or fine-tuned model as its large model. 
When \levi{} uses the original pre-trained model, the number of training parameters is significantly lower than all state-of-the-art baselines, and the number of inference parameters and FLOPs are comparable to those of a single large model. \levi{} using a fine-tuned model shows comparable results with single model-based baselines in all three metrics. Notably, compared to the heavy ensembles (i.e., FT+RobustModel, FT+FS), which consistently show good performances among the baselines in Tables~\ref{tbl:recom_main} and~\ref{tbl:vision_main}, \levi{} performs training and inference much faster while also achieving better OOD generalization.

\begin{table}[t]
  \caption{Number of parameters and FLOPs of baselines and \levi{} on T5x-small. The FLOPs of T5x is obtained from \citet{akbari-etal-2022-e}. See full results including all baselines and ViT in Sec.~\ref{appx:efficiency_all}.
  }
  \label{tbl:efficiency}
  \centering
  \scalebox{0.73}{
  \hspace{-0.4cm}
  \begin{tabular}{l@{\hspace{7pt}}c@{\hspace{5pt}}c@{\hspace{5pt}}c}
    \toprule
    %   &  \multicolumn{3}{c}{T5x-Small} \\
    Method & Params (Train.) & Params (Infer.) & FLOPs \\
    \cmidrule(lr){1-1}\cmidrule(lr){2-4}
    Fine-tuning & 60M & 60M & 33G \\
    \cmidrule(lr){1-1}\cmidrule(lr){2-4}
    LP$\rightarrow$FT~\citep{kumar2022finetuning} & 60M+$\sim$5K & 60M & 33G \\
    Baseline ensembles of two models & 120M & 120M & 66G \\
    Baseline ensembles of one model & 60M & 60M & 33G \\
    \cmidrule(lr){1-1}\cmidrule(lr){2-4}
    \levi{} with pre-trained model & $\sim$2M & 60M+$\sim$2M & 33G+$\sim$4M \\
    \levi{} with fine-tuned model & 60M+$\sim$2M & 60M+$\sim$2M & 33G+$\sim$4M \\
    \bottomrule
  \end{tabular}
  }
  \vspace{-0.4cm}
\end{table}

\subsection{Compatibility with Efficient Fine-Tuning Methods}
\label{sec:compatibility}

% We evaluate how \levi{} using a fine-tuned model performs when using LoRA to improve its training efficiency. 
We evaluate how \levi{} using a fine-tuned model performs with LoRA~\citep{hu2022lora} to improve its training efficiency.
In Table~\ref{tbl:lora}, \levi{} still improves the ID and OOD results of the LoRA-tuned models, indicating that \levi{} can be gracefully merged with existing efficient fine-tuning methods. 
% On vision datasets, \levi{} still improves the ID and OOD performances of the LoRA-tuned models (Table~\ref{tbl:lora}). Thus, \levi{} can be gracefully merged with existing efficient fine-tuning methods. 
We provide more discussion on possible extensions of \levi{} to further improve its efficiency via other efficient ensemble techniques (e.g., BatchEnsemble) in Sec.~\ref{appx:discussion}.
% When \levi{} uses a fine-tuned model instead of an original pre-trained model, we can use efficient fine-tuning techniques like LoRA~\citep{hu2022lora} to improve the overall training efficiency. Table~\ref{tbl:lora} shows that when we apply LoRA in the vision experiments, \levi{} can still improve the ID and OOD performances of the lora-tuned models. This result proves that \levi{} can be gracefully merged with existing efficient fine-tuning methods. 
% We also provide more discussion on possible extensions of \levi{} to further improve its efficiency using other efficient ensemble techniques (e.g., BatchEnsemble) in Sec.~\ref{appx:discussion}.

\begin{table}[t]
  \caption{LoRA results on the Medical and ImageNet datasets.
% \vspace{-0.2cm}
  }
  \label{tbl:lora}
  \centering
\scalebox{0.77}{
% \hspace{-0.3cm}
  \begin{tabular}{l@{\hspace{7pt}}c@{\hspace{5pt}}c@{\hspace{5pt}}c@{\hspace{5pt}}c}
    \toprule
      & \multicolumn{2}{c}{Medical} & \multicolumn{2}{c}{ImageNet} \\
    \cmidrule(lr){1-1}\cmidrule(lr){2-3}\cmidrule(lr){4-5}
      Method &  ID Acc & OOD Acc & ID Acc & OOD Acc \\
    \cmidrule(lr){1-1}\cmidrule(lr){2-3}\cmidrule(lr){4-5}
    LoRA-tuned ViT & 75.23\tiny{$\pm$0.01} & 73.06\tiny{$\pm$0.14} & \textbf{79.71\tiny{$\pm$0.09}} & 22.08\tiny{$\pm$0.17}\\
    \levi{} w.\@ LoRA-tuned ViT & \textbf{76.69\tiny{$\pm$0.17}} & \textbf{73.57\tiny{$\pm$0.06}} & 78.78\tiny{$\pm$0.21} & \textbf{25.64\tiny{$\pm$0.25}}\\
    \bottomrule
  \end{tabular}
  }
\vspace{-0.45cm}
\end{table}

% \vspace{-0.1cm}
\subsection{Ablation Study}
\label{sec:ablation}

We perform an ablation study on \levi{} to evaluate the impact of each component. 
%using the recommendation datasets. 
We first compare with simple final prediction ensembles (A1) between two fine-tuned large models and (A2) between one fine-tuned large model and one trained-from-scratch small yet task-specialized model, without using intermediate layers. We also compare with (A3) only ensembling intermediate layers without using the trained-from-scratch model. We then compare with solely using task-specialized models with (A4) single-head and (A5) multi-head -- see detailed settings in Sec.~\ref{appx:ablation_setting}.

% Table~\ref{tbl:ablation} shows the comparison with \levi{}, where the simple prediction ensembles (A1, A2) show both worse ID and OOD performances, the intermediate layer ensemble (A3) shows good results in OOD but not enough in ID, and the sorely using task-specialized models (A4, A5) achieve good results in ID but not enough in OOD.
Table~\ref{tbl:ablation} shows the comparison with \levi{}, where A1 and A2 show both worse ID and OOD results, A3 shows good results in OOD, but not enough in ID, and A4 and A5 achieve good results in ID but not enough in OOD.
We conclude that all components in \levi{} (i.e., using a small yet task-specialized model with multiple intermediate layers of a large model) contribute to the overall ID and OOD results.

\begin{table}[t]
  \caption{Ablation study on the recommendation tasks, where we consider genre shifts for MovieLens and time shifts for Amazon Review. We note that A1 and A2 are the final prediction ensembles, and A3 is the intermediate layer ensemble.
% \vspace{-0.2cm}
  }
  \label{tbl:ablation}
  \centering
\scalebox{0.75}{
% \hspace{-0.3cm}
%   \begin{tabular}{l@{\hspace{15pt}}c@{\hspace{9pt}}c@{\hspace{12pt}}c@{\hspace{9pt}}c}
  \begin{tabular}{l@{\hspace{7pt}}c@{\hspace{5pt}}c@{\hspace{5pt}}c@{\hspace{5pt}}c}
    \toprule
      & \multicolumn{2}{c}{MovieLens} & \multicolumn{2}{c}{Amazon Review} \\
    \cmidrule(lr){1-1}\cmidrule(lr){2-3}\cmidrule(lr){4-5}
      Method &  ID RMSE & OOD RMSE & ID RMSE & OOD RMSE \\
     \cmidrule(lr){1-1}\cmidrule(lr){2-3}\cmidrule(lr){4-5}
    T5x Fine-tuning  & 1.175\tiny{$\pm$0.021} & 1.268\tiny{$\pm$0.036} & 1.386\tiny{$\pm$0.024} & 1.543\tiny{$\pm$0.014} \\
    \cmidrule(lr){1-1}\cmidrule(lr){2-3}\cmidrule(lr){4-5}
    A1) T5x+T5x  & 1.103\tiny{$\pm$0.011} & 1.178\tiny{$\pm$0.021} & 1.362\tiny{$\pm$0.013} & 1.531\tiny{$\pm$0.010}\\
    A2) T5x+MLP  & 1.035\tiny{$\pm$0.009} & 1.163\tiny{$\pm$0.015} & 1.272\tiny{$\pm$0.008} & 1.455\tiny{$\pm$0.019}\\
    A3) T5x Intermediates & 1.125\tiny{$\pm$0.003} & 1.142\tiny{$\pm$0.012} & 1.137\tiny{$\pm$0.001} & \textbf{1.285\tiny{$\pm$0.007}}\\
    A4) Single-head MLP & 1.003\tiny{$\pm$0.013} & 1.158\tiny{$\pm$0.101} & 1.104\tiny{$\pm$0.005} & 1.326\tiny{$\pm$0.002}\\
    A5) Multi-head MLP & 1.001\tiny{$\pm$0.020} & 1.121\tiny{$\pm$0.034} & 1.102\tiny{$\pm$0.004} & 1.319\tiny{$\pm$0.036}\\
    \cmidrule(lr){1-1}\cmidrule(lr){2-3}\cmidrule(lr){4-5}
    \levi{} & \textbf{0.932\tiny{$\pm$0.005}} & \textbf{1.065\tiny{$\pm$0.018}} & \textbf{1.095\tiny{$\pm$0.003}} & 1.310\tiny{$\pm$0.006}\\
    \bottomrule
  \end{tabular}
  }
  \vspace{-0.3cm}
\end{table}

\subsection{Effects of Different Intermediate Layers}
\label{sec:algorithm_analysis}
% \vspace{-0.1cm}

\begin{figure}[t]
% \vspace{-0.1cm}
\centering
\includegraphics[width=\columnwidth,trim=0cm 0.3cm 0cm 0cm]{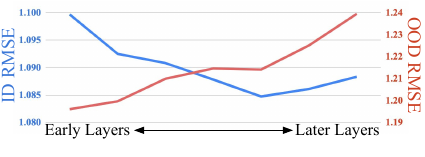}
\vspace{-0.5cm}
\caption{\small Effects of intermediate layers on ID (blue) and OOD (red) performances, where lower RMSE is better. We report the average results on MovieLens and Amazon Review using T5x.}
\vspace{-0.2cm}
\label{fig:intermediate}
\end{figure}

We investigate the roles of different intermediate layers used in \levi{}.
% To further understand the performance of \levi{}, we analyze the effects of different intermediate layers in pre-trained models. 
We use one intermediate layer at each time and attach a small MLP classification module to the target intermediate layer for training.
Interestingly, Fig.~\ref{fig:intermediate} shows that the later and early layers tend to be more useful for ID and OOD performances, respectively. This result is consistent with previous work that early layers are more general (robust), and the later layers are more specific (accurate)~\citep{yosinski2014transferable}. A recent study also observes that using multiple intermediate layers can improve the overall model robustness compared to simple linear probing~\citep{evci2022head2toe}. Similarly, this result demonstrates that using both the later and early layers can improve the ID-OOD tradeoffs.

\subsection{\yuji{Discussion on the distribution gap (domain gap) between pre-training, fine-tuning, and test data}}

\yuji{We now discuss how the distribution gap (domain gap) between pre-training, fine-tuning, and test data affects LEVI. Revisiting our results in Tables~\ref{tbl:recom_main} and~\ref{tbl:vision_main}, we observe different degrees of domain gap. First, Table~\ref{tbl:recom_main} shows scenarios where the pre-training domain is relatively distant from the downstream fine-tuning and test domains (i.e., using general language pre-trained models in recommendation domains). On the other hand, Table~\ref{tbl:vision_main} covers cases where the pre-training domain is more similar to the downstream domains (i.e., using image pre-trained models in other image domains). Across both scenarios, LEVI shows clear performance improvements over baseline methods.}

\yuji{In general, we anticipate that LEVI performs especially well when the pre-training and downstream domains differ (e.g., language vs. recommendation domains), as LEVI's advantage comes from breaking the previous assumption that the pre-training domain already has good enough features.}

\vspace{-0.1cm}
\section{Conclusion}
% \vspace{-0.1cm}

% We identified and addressed key limitations in both pre-trained models and fine-tuning data that can hurt the out-of-distribution generalization of fine-tuned models. 
We proposed a novel fine-tuning method \levi{}, which tightly ensembles a large pre-trained model with a smaller task-specific model for improving OOD generalization of fine-tuning.
We first identified that the inherent issues in both pre-trained models and fine-tuning data can jointly hurt the OOD generalization of fine-tuned models.
To address these issues, \levi{} combines two complementing models to suppress their problems while preserving useful features for downstream tasks, leading to improved OOD generalization, especially when pre-training and downstream distributions are largely different. Experiments on large language and vision models showed that \levi{} greatly enhances fine-tuning OOD generalization while not losing ID performances.
We believe \levi{} reveals the value of using trained-from-scratch models in mitigating the limitations of pre-trained models.

\section*{Impact Statement}

We believe our work on improving fine-tuning generalization can positively impact society by making AI models being more robust and safe when deployed in real-world applications. Specifically, generalizable AI models would be less prone to failures or unpredictable behaviors when faced with new inputs from unseen distributions, leading to increased reliability and safety. Also, AI models with a strong generalization ability can improve people's lives in areas that require high adaptability across diverse contexts, including healthcare, finance, and recommendation systems.

We do note that as our framework is based on supervised learning that requires training (fine-tuning) data, considering privacy and fairness issues in the training data will become essential. Many recent studies reveal that machine learning models, especially large models applied irresponsibly, could amplify privacy concerns and societal unfairness~\citep{Bommasani2021FoundationModels,barocas2023fairness}. Thus, one needs to carefully build the training data for \levi{} to prevent potential negative impacts, especially when the target applications highly affect society.

% \section*{Acknowledgements}
% \scalebox{0.98}{This work was partly supported by a Google Research Award.}

% \bibliography{main}
% \bibliographystyle{iclr2024_conference}

\bibliography{main}
\bibliographystyle{icml2024}

\newpage
\appendix
\onecolumn

% \vspace{0.2cm}
\section{More Related Work}
\label{appx:relatedwork}

Continuing from Sec.~\ref{sec:relatedwork}, we discuss more related work on traditional studies of out-of-distribution (OOD) generalization and other related fields (e.g., model robustness for noisy or adversarial data, domain adaptation, and OOD detection).

\paragraph{OOD Generalization in the Traditional Literature} Among the broad model generalization issues, making the model more robust to various OOD scenarios becomes indispensable for AI deployment~\citep{shen2021towards}. The main approaches for OOD generalization can be categorized into: 1) unsupervised representation learning that finds a better representation for diverse distributions~\citep{mahajan2021domain, zhang2022towards, harary2022unsupervised, 9088148},
% ~\citep{mahajan2021domain, zhang2022towards, harary2022unsupervised, 9088148}, 
2) supervised model training that modifies model architectures or training processes to prevent the model from losing the generalization~\citep{finn2017model,li2018learning,d2019domain,carlucci2019domain, raghunathan2020understanding, xie2020n}, and 3) specialized optimization designing that uses robustness-aware objectives to ensure OOD performances (e.g., distributionally robust optimization and invariant risk minimization)~\citep{Sinha2017CertifyingSD,arjovsky2019invariant}.

Among the above approaches, the supervised model training approaches are the most relevant category to ours, as they design new training processes by using given labeled training data. 
% -- see detailed discussion for other approaches in Appx.~\ref{appx:relatedwork}. 
For example, a traditional approach is to utilize meta learning~\citep{finn2017model,li2018learning}, where a model trained on a variety of tasks adapts to the new tasks using a small number of training samples. 
Alternatively, adversarial training-based approaches~\citep{roh2020fr, pmlr-v139-yi21a} train a model with perturbed input data or additional discriminator to make the model robust to OOD data.
Also, several self-training-based approaches~\citep{raghunathan2020understanding, xie2020n} show promising improvements in generalization by leveraging additional unlabeled OOD data.
However, many traditional studies do not consider the challenges in large models (e.g., inherent problems in the pre-trained features, scalability) or assume to access pre-training or unlabeled test data. In comparison, our work improves generalization in the fine-tuning pre-trained model paradigm, where 1) the given models may already have inherent issues, 2) the training and inference efficiency becomes more important, and 3) additional information on the pre-training or deployment (OOD) data is usually unavailable.

\paragraph{Other Related Work}
Although not our immediate focus, there are other noteworthy directions, including 1) model robustness for noisy or adversarial data, which aims to maintain model accuracy even when the input data is corrupted~\citep{song2022learning, gowal2021improving, shen2019learning, carmon2019unlabeled, menon2019can, ren2018learning}, 2) domain adaptation, which focuses on enabling a model trained on one domain to perform effectively on different yet related domains~\citep{wang2018deep, wang2022continual, wang2021tent, li2020model, kundu2020universal, tzeng2017adversarial, ganin2015unsupervised}, and 3) OOD detection, which concentrates on identifying data that are not from the training distribution~\citep{yang2021generalized, fort2021exploring, liu2020energy, winkens2020contrastive, ren2019likelihood, vyas2018out}. In comparison, we focus on training generalizable models in naturally occurring OODs, but do not explicitly consider adversarial scenarios or OOD detection. Extending \levi{} to support these directions can be an interesting future work.

\yuji{Finally, we discuss other related studies that use relevant ideas with some parts of \levi{}'s architecture.
First, \citet{li-liang-2021-prefix} and \citet{houlsby2019parameter} use adapting layers (adaptors) in the model architecture for achieving parameter-efficient fine-tuning, where the adaptors are used to improve full fine-tuning performances with fewer parameters. In contrast, we utilize the adapting layers to merge complementing information from two very different models for OOD generalization. 
Also, \citet{lee2023surgical} and \citet{shen2021partial} use partial fine-tuning strategies in the OOD setting, but they require small amounts of labeled new domain data (i.e., the OOD test data). In comparison, LEVI does not assume any prior knowledge of OOD data.
In addition, \citet{zoph2020rethinking}, \citet{he2019rethinking}, and \citet{shen2017dsod} show that solely using the trained-from-scratch models can perform better than using fine-tuned pre-training models in object detection and segmentation.
In contrast, we jointly use both pre-trained and trained-from-scratch models to utilize their strengths while mitigating their own inherent problems. 
}

\vspace{0.4cm}
% \newpage
\section{Synthetic Example}
\label{appx:synthetic_ex}

Continuing from Sec.~\ref{sec:theory}, we provide a synthetic example illustrating that fine-tuned models can have a worse generalization ability compared to pre-trained and trained-from-scratch models.

Let a target task consists of five features [$x_1, x_2, x_3, x_4, x_5$], where $x_1$ and $x_5$ are spurious features, and $x_2$, $x_3$, and $x_4$ are transferable features. In the following scenario, $x_1$ and $x_5$ are originated from pre-trained features and fine-tuning data, respectively. Here are the settings:
\begin{itemize}[topsep=0ex,partopsep=0ex,leftmargin=7mm,itemsep=0ex]
    % \item Let a target task consists of five features [$x_1, x_2, x_3, x_4, x_5$], where $x_1$ and $x_5$ are spurious features, and $x_2$, $x_3$, and $x_4$ are transferable features. In the following example, $x_1$ and $x_5$ are originated from pre-trained features and fine-tuning data, respectively.
    \item We consider a linear model $\ry = \vw\rx^T$. Here, the optimal model weights $\vw_\text{true}$ are $[0, 1, 1, 1, 0]$, which do not rely on the spurious features $x_1$ and $x_5$ while only using transferable features.
    \item \textit{[Fine-tuning data]} We have three fine-tuning data points $(\rx^{(1)}, \ry^{(1)}) = ([0, 0, \frac{1}{3}, \frac{1}{3}, \frac{1}{3}], 1)$, $(\rx^{(2)}, \ry^{(2)}) = ([0, 0, -\frac{1}{2}, -\frac{1}{2}, 0], -1)$, and $(\rx^{(3)}, \ry^{(3)}) = ([0, 0, \frac{1}{2}, \frac{1}{4}, \frac{1}{4}], 1)$, which are affected by the spurious feature $x_5$.
    \item \textit{[Pre-trained model]} Here, we assume a pre-trained model with weights $\vw_\text{pretrain} = [1, 1, 1, 0, 0]$, which utilizes the spurious feature $x_1$.
    \item \textit{[Fine-tuned model]} Here, when we fine-tune the given pre-trained model with the above two data points, the fine-tuned model weights become $\vw_\text{finetune} = [1, 1, 1, 1, 1]$.
    \item \textit{[Trained-from-scratch model]} On the other hand, when we train a model from scratch, the model weights can be $\vw_\text{train-from-scratch} = [0, 0, 1, 1, 1]$.
\end{itemize}

As a result, we have three models $\vw_\text{pretrain} = [1, 1, 1, 0, 0]$, $\vw_\text{train-from-scratch} = [0, 0, 1, 1, 1]$, and $\vw_\text{finetune} = [1, 1, 1, 1, 1]$. 

Table~\ref{tbl:synthetic_example} shows four test data points, where there is no correlation between the label $\ry$ and spurious features $x_1$ and $x_5$. We note that the optimal model with $\vw_\text{true}=[0,1,1,1,0]$ can predict all test samples correctly with zero errors. When we apply the three models $\vw_\text{pretrain}$, $\vw_\text{train-from-scratch}$, and $\vw_\text{finetune}$ to these test data points, we get the predictions $\hat{\ry}$ as in the fourth column in Table~\ref{tbl:synthetic_example}. As the fine-tuned model $\vw_\text{finetune}$ uses both spurious features $x_1$ and $x_5$ while $\vw_\text{pretrain}$ and $\vw_\text{train-from-scratch}$ use either $x_1$ or $x_5$, the errors of the fine-tuned model are worse than the other two models' errors (last column in Table~\ref{tbl:synthetic_example}). This observation is consistent with our previous discussions, including Corollary~\ref{corollary:finetuned} and the results in Figures~\ref{fig:overview} \&~\ref{fig:toy_example}.

% [[Add test data samples and error values]].
% $(\rx^{(1)}_\text{test}, \ry^{(1)}_\text{test}) = ([1, \frac{1}{3}, \frac{1}{3}, \frac{1}{3}, 1], 1)$, $(\rx^{(2)}_\text{test}, \ry^{(2)}_\text{test}) = ([1, -\frac{1}{3}, -\frac{1}{3}, -\frac{1}{3}, 1], -1)$, $(\rx^{(3)}_\text{test}, \ry^{(3)}_\text{test}) = ([-1, \frac{1}{3}, \frac{1}{3}, \frac{1}{3}, -1], 1)$, and $(\rx^{(4)}_\text{test}, \ry^{(4)}_\text{test}) = ([-1, -\frac{1}{3}, -\frac{1}{3}, -\frac{1}{3}, -1], -1)$

\begin{table}[h]
  \caption{Test data points, model predictions, and errors.
  }
  \label{tbl:synthetic_example}
  \centering
  \scalebox{0.95}{
  \begin{tabular}{ccccc}
    \toprule
    $\rx_\text{test}$ & $\ry_\text{test}$ & Model & $\hat{\ry}$ & L1 loss (i.e, $|\ry_\text{test}-\hat{\ry}|$)\\
    \midrule
    \multirow{3}{*}{$[1, \frac{1}{3}, \frac{1}{3}, \frac{1}{3}, 1]$} & \multirow{3}{*}{1} & $\vw_\text{pretrain}$ & 5/3 & 2/3\\
    & & $\vw_\text{train-from-scratch}$ & 5/3 & 2/3\\
    & & $\vw_\text{finetune}$ & 3 & 2\\
    \midrule
    \multirow{3}{*}{$[1, -\frac{1}{3}, -\frac{1}{3}, -\frac{1}{3}, 1]$} & \multirow{3}{*}{-1} & $\vw_\text{pretrain}$ & 1/3 & 4/3\\
    & & $\vw_\text{train-from-scratch}$ & 1/3 & 4/3\\
    & & $\vw_\text{finetune}$ & 1 & 2\\
    \midrule
    \multirow{3}{*}{$[-1, \frac{1}{3}, \frac{1}{3}, \frac{1}{3}, -1]$} & \multirow{3}{*}{1} & $\vw_\text{pretrain}$ & -1/3 & 4/3 \\
    & & $\vw_\text{train-from-scratch}$ & -1/3 & 4/3\\
    & & $\vw_\text{finetune}$ & -1 & 2\\
    \midrule
    \multirow{3}{*}{$[-1, -\frac{1}{3}, -\frac{1}{3}, -\frac{1}{3}, -1]$} & \multirow{3}{*}{-1} & $\vw_\text{pretrain}$ & -5/3 & 2/3\\
    & & $\vw_\text{train-from-scratch}$ & -5/3 & 2/3\\
    & & $\vw_\text{finetune}$ & -3 & 2\\
    \bottomrule
  \end{tabular}
  }
%   \vspace{-0.1cm}
\end{table}

\section{Weighted Sum of Loss Functions}
\label{appx:weight_sum}

\setlength{\abovedisplayskip}{0.3cm}
\setlength{\belowdisplayskip}{0.3cm}

Continuing from Sec.~\ref{sec:framework}, we discuss a possible extension of the loss function in our framework. 

In Sec.~\ref{sec:framework}, we introduce \levi{}'s loss function as follows:
\begin{gather*}
\min_{\vw} \frac{1}{m} \cdot \frac{1}{n} \sum^{m}_{i} \sum^{n}_{j} \ell(\ry_i,\hat{\ry}^{(j)}_i),
\end{gather*}
where $m$ and $n$ are the numbers of training samples and adapting layers, respectively, and $\hat{\ry}^{(j)}_i$ is the predicted label from the adapting layer $j$ for the input data $i$ -- see the architecture of \levi{} in Figure~\ref{fig:genie}, which illustrates each component in the loss function.
We update \levi{} by using the above loss function, which allows us to tightly ensemble the information from both pre-trained and trained-from-scratch (i.e., randomly initialized then trained) models in the adapting layers.

We basically consider the equal weights on all $\hat{\ry}^{(j)}$, but one can change it as a weighted sum of $\hat{\ry}^{(j)}$ as follows:
\begin{gather*}
\min_{\vw} \frac{1}{m} \cdot \frac{1}{n} \sum^{m}_{i} \sum^{n}_{j} w^{(j)}\ell(\ry_i,\hat{\ry}^{(j)}_i)\\
\text{s.t.} \sum^{n}_{j} w^{(j)} = 1,~~ w^{(j)} \geq 0,
\end{gather*}
where $w^{(j)}$ is the weight for the loss of each adapting layer $j$ (i.e., $\ell(\ry,\hat{\ry}^{(j)})$). Here, $w^{(j)}$ can be a hyperparameter to tune the importance between different intermediate layers. For example, when we aim to focus on more specific (general) features, we can give more weight to the later (early) layers.

\vspace{0.5cm}
\section{Experimental Settings}
\label{appx:settings}

Continuing from Sec.~\ref{sec:experiments}, we provide more details on experimental settings. In all experiments, we use Dragonfish TPU (i.e., TPUv3) and Jellyfish TPU (i.e., TPUv2) with 2x2 topology for T5x and ViT experiments, respectively. Also, as we discussed in Sec.~\ref{sec:experiments}, we use TensorFlow~\citep{tensorflow2015-whitepaper} with JAX~\citep{jax2018github} and Flax~\citep{flax2020github}.

\subsection{Datasets and Pre-processings}
\label{appx:preprocessing}

We consider four datasets: MovieLens~\citep{10.1145/2827872}, Amazon Review~\citep{ni2019justifying}, Diabetic Retinopathy (Medical)~\citep{diabetic-retinopathy-detection}, and ImageNet-Variants~\citep{wang2019learning, hendrycks2021many, hendrycks2021nae, recht2019imagenet}. In this paper, we use MovieLens and Amazon Review for language-based recommendation tasks and Medical and ImageNet-Variants for computer vision tasks. All datasets are from the TensorFlow Datasets library~\citep{TFDS}.

% We first provide data examples for each dataset. 
Tables~\ref{tbl:movielens_example} and~\ref{tbl:amazonreview_example} show the data examples of the MovieLens and Amazon Review datasets.
\begin{itemize}[topsep=0ex,partopsep=0ex,leftmargin=7mm,itemsep=0ex]
\item MovieLens~\citep{10.1145/2827872} contains movie rating data from online movie website, and we utilize user-id, user-age, user-occupation, user-zipcode, movie-title, and movie-id as input features and rating as the label attribute, where the rating range is $[1, 5]$. We use a stable version of MovieLens that contains 100,000 ratings from 943 users on 1,682 movies. Each user gives ratings on at least 20 movies. For the OOD scenario, we consider \textit{genre (subpopulation) shifts}, where the ID data contains movies of top-5 popular genres (action, comedy, drama, romance, thriller), and the OOD data contains other 12 genres (e.g., animation, sci-fi) that have at least 200 data points. We construct the ID and OOD datasets to be mutually exclusive.
\item Amazon Review~\citep{ni2019justifying} contains product rating data from the Amazon.com website, and we utilize customer-id, product-title, and product-id as input features and rating as the label attribute, where the rating range is $[1, 5]$. For the OOD scenarios, we consider \textit{time shifts} and \textit{product (subpopulation) shifts}. Here, the ID data contains the first 4 years' (oldest) ratings of books, and the OOD data contains the most recent year's ratings of books (i.e., time shifts) and other products (i.e., product shifts), including watch, toy, sports, music, jewelry, furniture, and baby.
\item When we serve these tabular-based data into the language model, we use the method proposed in~\citet{dinh2022lift}, which concatenates all attribute values into one sentence. For example, the first row in Table~\ref{tbl:movielens_example} can be converted into an input sentence \textit{``When the user id is 138, the user occupation is doctor, the user zipcode is 53211, the movie title is One Flew Over the Cuckoo's Nest (1975), and the movie id is 357, what can be the user rating on this movie?:''}. Also, when we serve the data into the small MLP model of \levi{}, we pre-process the data so that it can be used as the input for the embedding layers. Here, we follow the pre-processing method used in \citet{10.1145/3523227.3546767}. 
\end{itemize}

\begin{table}[h]
\vspace{-0.3cm}
  \caption{MovieLens data examples.
  }
  \label{tbl:movielens_example}
  \centering
  \scalebox{0.9}{
  \begin{tabular}{ccccccc}
    \toprule
    movie genres & user id & user occupation & user zipcode & movie title & movie id & rating\\
    % movie genres & user id & user occupation & user zipcode & user age & user gender & movie title & movie id & rating\\
    \midrule
    % Drama & 138 & doctor & 53211 & 46 & True & One Flew Over the Cuckoo's Nest (1975) & 357 & 4\\
    Drama & 138 & doctor & 53211 & One Flew Over the Cuckoo's Nest (1975) & 357 & 4\\
    Comedy, Romance & 92 & entertainment & 80525 & Strictly Ballroom (1992) & 709 & 2\\
    Comedy & 301 & student & 55439 & Very Brady Sequel, A (1996) & 412 & 4\\ 
    Crime, Drama & 60 & healthcare & 06472 & Pulp Fiction (1994) & 56 & 4\\
    Horror, Thriller & 197 & technician & 75094 & Scream 2 (1997) & 895 & 3\\
    Drama, Thriller &  601 & artist & 99687 & Crash (1996) & 325 & 4 \\
    Animation, Children, ... & 710 & student & 92020 & Aladdin (1992)& 95 & 3\\
    Action, Crime, Romance & 833 & writer & 90019 & True Romance (1993) & 92 & 2\\
    Comedy & 916 & engineer & N2L5N & Bob Roberts (1992) & 425 & 5 \\
    Adventure, Sci-Fi, ... & 920 & administrator & 02215 & Starship Troopers (1997) & 271 & 2\\
    \bottomrule
  \end{tabular}
  }
%   \vspace{-0.1cm}
\end{table}

\begin{table}[h]
  \caption{Amazon Review data examples. We do not use the review body and title, which strongly indicate the ratings.
  }
  \label{tbl:amazonreview_example}
  \centering
  \scalebox{0.9}{
  \begin{tabular}{cccccc}
    \toprule
    product category & review date & customer id & product title & product id & rating\\
    \midrule
    Books & 2000-08-10 & 51389465 & Admission Of Love (Arabesque) & 1583141642 & 5\\
    Books & 2005-10-04 & 23641112 & Solve Your Child's Sleep Problems & 0671620991 & 4\\
    Watches & 2013-08-20 & 45902750 & Timex Men's Expedition Metal Field Watch & B004VRD6FY & 3 \\
    Toys & 2014-07-16 & 28742093 & Ticket To Ride - Europe & B000809OAO & 4\\
    Sports & 2015-08-29 & 4005801 & Chipolo Bluetooth Item Tracker (Blue) & B00L177Z3Q & 1\\
    Music & 2009-04-27 & 10801536 & A Mi Edad & B001P0XNJ4 & 5\\
    Jewelry & 2015-02-05 & 18490528 & Sterling Silver Twisted Love Knot Stud Earrings & B00CRMQ8OG & 4\\
    Furniture & 2013-03-03 & 28467583 & Rust Orange Full Sized with Arms Convertible Sofa & B0070ZERCA & 3\\ 
    Baby & 2014-10-15 & 20757152 & Skip Hop Baby Duo Signature Diaper Bag & B00J4J2AQK & 5\\
    \bottomrule
  \end{tabular}
  }
  \vspace{0.2cm}
\end{table}

\begin{figure}[h]
\vspace{-0.2cm}
\centering
\includegraphics[width=0.52\columnwidth,trim=0cm 0.3cm 0cm 0cm]{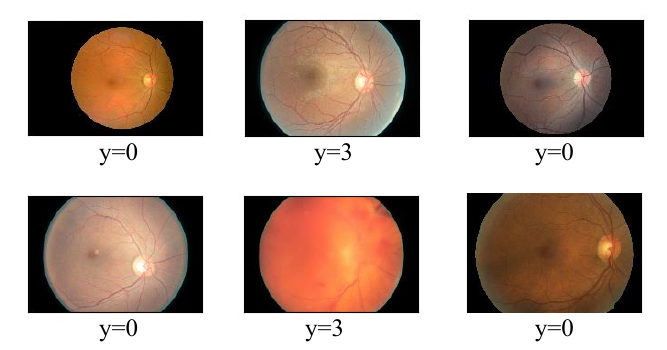}
\vspace{-0.1cm}
\caption{\small Diabetic Retinopathy (Medical) data examples for the in-distribution. The images are from TensorFlow Datasets~\citep{TFDS}.}
\vspace{-0.4cm}
\label{fig:medical_training}
\end{figure}

\begin{figure}[h]
% \vspace{-0.1cm}
\centering
\includegraphics[width=0.5\columnwidth,trim=0cm 0.3cm 0cm 0cm]{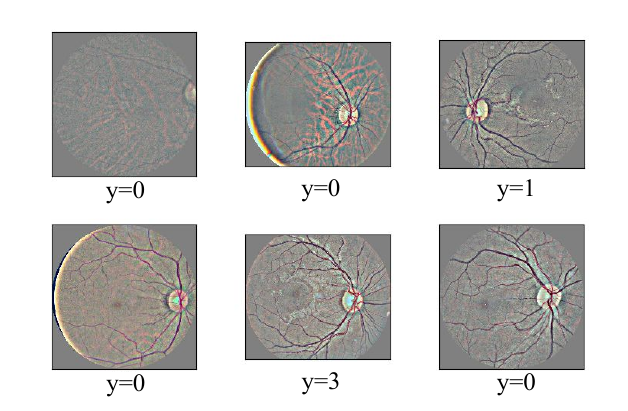}
\vspace{-0.1cm}
\caption{\small Diabetic Retinopathy (Medical) data examples for the out-of-distribution. The images are from TensorFlow Datasets~\citep{TFDS}.}
% \vspace{-0.4cm}
\label{fig:medical_test}
\end{figure}

\begin{figure}[h]
\vspace{0.7cm}
\centering
\includegraphics[width=0.63\columnwidth,trim=0cm 0.3cm 0cm 0cm]{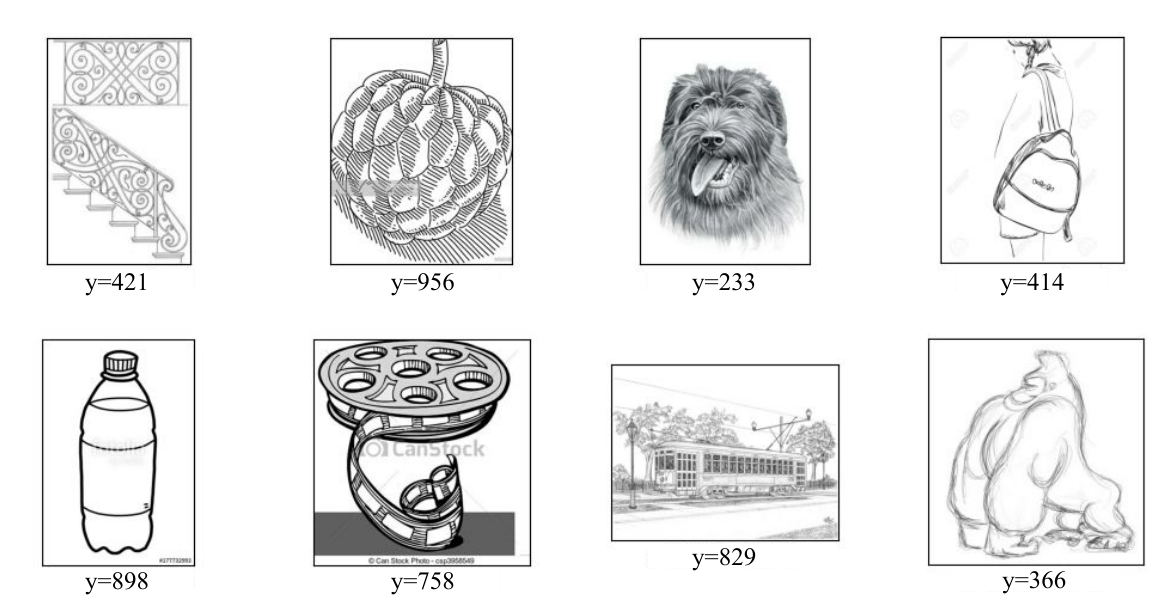}
\vspace{-0.1cm}
\caption{\small ImageNet-Sketch data examples used for the in-distribution of ImageNet experiments. The images are from TensorFlow Datasets~\citep{TFDS}.}
% \vspace{-0.3cm}
\label{fig:imagenet_training}
\end{figure}

\begin{figure}[h]
% \vspace{-0.1cm}
\centering
\includegraphics[width=0.8\columnwidth,trim=0cm 0.3cm 0cm 0cm]{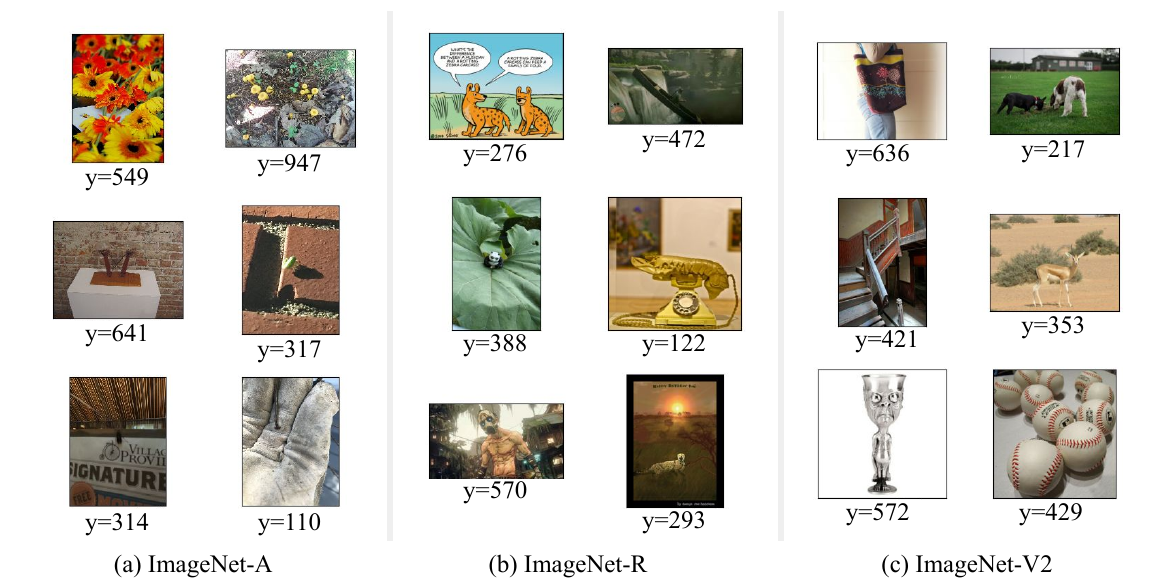}
\vspace{-0.1cm}
\caption{\small ImageNet-A (a), ImageNet-R (b), and ImageNet-V2 (c) data examples used for out-of-distributions of ImageNet experiments. The images are from TensorFlow Datasets~\citep{TFDS}.}
% \vspace{-0.5cm}
\label{fig:imagenet_test}
\end{figure}

% \newpage
Figures~\ref{fig:medical_training} \&~\ref{fig:medical_test} and Figures~\ref{fig:imagenet_training} \&~\ref{fig:imagenet_test} show the data examples of the Diabetic Retinopathy (Medical) and ImageNet-Variants datasets, respectively.
\begin{itemize}[topsep=0ex,partopsep=0ex,leftmargin=7mm,itemsep=0ex]
\item Diabetic Retinopathy (Medical)~\citep{diabetic-retinopathy-detection} contains human retina images, where the label attribute is the severity of diabetic retinopathy in the range $[0, 4]$. For the OOD scenario, we consider \textit{quality shifts}: the ID data (Figure~\ref{fig:medical_training}) and OOD data (Figure~\ref{fig:medical_test}) show different image styles and resolutions. 
\item ImageNet-Variants~\citep{wang2019learning, hendrycks2021many, hendrycks2021nae, recht2019imagenet} contain different styles (e.g., sketch, adversarial) of ImageNet datasets with 1,000 label classes. For the OOD scenario, we consider \textit{domain shifts}: the ID dataset is ImageNet-Sketch (Figure~\ref{fig:imagenet_training}), and the OOD datasets are ImageNet-A, ImageNet-R, and ImageNet-V2 (Figure~\ref{fig:imagenet_test}). ImageNet-Sketch consists of about 50,000 black and white sketch images, where each ImageNet class has 50 images. All these datasets share the original ImageNet classes. ImageNet-Sketch~\citep{wang2019learning} consists of about 50,000 black and white sketch images, where each class contains 50 images. ImageNet-A~\citep{hendrycks2021nae} consists of adversarial images that are wrongly classified by ResNet-50 models. ImageNet-R~\citep{hendrycks2021many} consists of rendition images, which contain art, cartoons, graffiti, tattoos, toys, video games, and other renditions. ImageNet-V2~\citep{recht2019imagenet} consists of new test data for ImageNet, which are collected a decade after the original ImageNet dataset. 
\yuji{Previous works~\citep{kumar2022finetuning, wortsman2022robust} used an ImageNet pre-trained model, fine-tuned it with a smaller version of ImageNet, and tested it on other ImageNet variants (e.g., ImageNet-A and ImageNet-R). Compared to this setting, we focus on a more challenging scenario where the fine-tuning data is different from the pre-training data; we thus fine-tune the model on ImageNet-Sketch, which has a different image style from the original ImageNet, and test on other variants (ImageNet-A, ImageNet-R, and ImageNet-V2).}
\item We resize all images into 224$\times$224 pixels.
\end{itemize}

% AmazonBooks: customer_id, product_id, product_title
% MovieLens: 
% "inputs": vocabulary.encode(text[:-1]),
% "targets": vocabulary.encode(text[-1:]),
% "movie_genres": movie_genres,
% "movie_title": movie_title,
% "movie_id": tf.expand_dims(movie_id, axis=0),
% "user_id": tf.expand_dims(user_id, axis=0),
% "raw_user_age": tf.expand_dims(  # New
%     tf.convert_to_tensor(d["raw_user_age"]), axis=0
% ),
% "user_occupation_label": tf.expand_dims(  # New
%     tf.convert_to_tensor(d["user_occupation_label"]), axis=0
% ),
% "user_zip_code": tf.expand_dims(user_zip_code, axis=0),  # New
% "user_rating": tf.expand_dims(
%     tf.convert_to_tensor(d["user_rating"]), axis=0
% ),

% \vspace{0.5cm}
\newpage
\subsection{Models}
\label{appx:model_settings}

% Model architecture details on both T5x, ViT, and our ensembles (MLP, embeddings, and CNN).

In our experiments, we use two large pre-trained models: T5x~\citep{2020t5, roberts2022t5x} and ImageNet-21k pre-trained ViT~\citep{dosovitskiy2020vit} for language-based recommendation and computer vision tasks, respectively. Specifically, we use T5x-small and ViT-base architectures, which have 60M and 86M parameters, respectively. We modify the last layers of T5x and ViT to work with each classification task.

\levi{} uses a small randomly-initialized model and adapting layers together with the given large model (i.e., T5x and ViT). We note that the small randomly-initialized model in \levi{} can be defined to effectively learn the task-specialized features, and we follow the general knowledge from recommendation and computer vision literature.
\begin{itemize}[topsep=0ex,partopsep=0ex,leftmargin=5mm,itemsep=0ex]
\item For the large model, \levi{} can use either an original pre-trained model or an adapted (fine-tuned or light-tuned) model, as discussed in Remark~\ref{remark:using_adapted}. In our main experiments, we perform \levi{} with a light-tuned model that updates half of the parameters (i.e., half of the transformers) to improve the overall performance by aligning a given large model with the target task. We also provide the full results of \levi{} with pre-trained, light-tuned, and fine-tuned models in Sec.~\ref{appx:different_largemodels}.
\item For the small randomly-initizlied model, we use a two-layer multi-layer perceptron (MLP) model with input embedding layers for recommendation tasks and a four-layer convolutional neural network (CNN) model for computer vision tasks. 
The two-layer MLP model used in the recommendation tasks consist of 512 and 256 neurons in each hidden layer. The CNN model used in the vision tasks consist of four convolution layers ((3$\times$3)-kernel, 32 features), ((3$\times$3)-kernel, 64 features), ((3$\times$3)-kernel, 64 features), and ((3$\times$3)-kernel, 64 features), where a batch normalization, ReLU, and max pooling are followed by each convolution layer. The last layer is flattened to be served to the adapting layers.
\yuji{We note that, in general, as the small model size increases (e.g., it has more layers), LEVI's performance increases, but eventually converges. In our experiments, the small models with good enough performances are much smaller than any large pre-trained models used in the language-based and vision-based tasks.}
\item Each adapting layer is composed of an MLP with one hidden layer, where the numbers of neurons in the hidden layer are 256 and 1024 for the recommendation and computer vision tasks, respectively.
\end{itemize}

\vspace{0.5cm}
\subsection{Hyperparameters}
\label{appx:hyperparameters}

We provide more details on hyperparameters and settings. 

Here are common hyperparameters and settings for all algorithms. 
We use the Adam optimizer and SGD optimizer for T5x and ViT experiments, respectively. 
For batch sizes, we use 200 for MovieLens, 100 for Amazon Review, and 512 for all computer vision datasets.
For learning rates, we consider a set \{0.0001, 0.001, 0.01, 0.1\} for all algorithms except linear probing. In linear probing, we use a learning rate set with larger values \{0.001, 0.01, 0.1, 1.0, 10.0\}, as reported in previous studies~\citep{kumar2022finetuning,he2020momentum,chen2020generative}. For each algorithm, we choose the best hyperparameters from the above candidate sets to achieve the best performance in the in-distribution validation set, while not accessing the out-of-distribution datasets.

When the baselines require other hyperparameters (e.g., FT+ZeroShot~\citep{wortsman2022robust} uses a weight parameter to balance the importance between the fine-tuned model and the zero-shot model), we follow the candidate sets used in the original paper and choose the best hyperparameters to achieve the best performance in the in-distribution validation set.

\vspace{0.5cm}
\subsection{Settings of Ablation Study}
\label{appx:ablation_setting}

We provide more details on the setting of the ablation study.
We consider five ablations (A1--A5), as shown in Table~\ref{tbl:ablation}. First, we perform a simple prediction ensemble between two fine-tuned large models (A1). We also perform a simple prediction ensemble between a fine-tuned large model and a small trained-from-scratch model (A2). We note that both A1 and A2 do not use intermediate layer information and simply average the final outputs (predictions) of the models. We then perform an ensemble between intermediate layers of a large model without using the trained-from-scratch model (A3). Finally, we compare with trained-from-scratch (task-specialized) models with (A4) single-head and (A5) multi-head without using the large model. Here, A4 is a standard model that produces one prediction per input sample. A5 consists of a shared-bottom layer block followed by multiple final heads that produce predictions, where we average the predictions of the different heads, similar to the ensemble.

Based on the above ablations, we investigate the importance of each component in \levi{}. For example, A1 and A3 show the impact of using the small yet task-specialized trained-from-scratch model in \levi{}. Also, A2, A4, and A5 show the importance of using the large model. Finally, A1 and A2 show the benefits of using the intermediate layer outputs in the large model. As a result, Table~\ref{tbl:ablation} show that all components in \levi{} (i.e., using a small yet task-specialized model with multiple intermediate layers of a large model) contribute to the overall ID and OOD performances.

% \newpage
\vspace{0.5cm}
\section{Additional Experiments}

Continuing from Sec.~\ref{sec:experiments}, we provide additional experimental results.

\subsection{More Results on Recommendation Tasks}
\label{appx:recom_all}

Continuing from Sec.~\ref{sec:performances_id_ood}, we provide full OOD performances on the MovieLens genre shifts in Tables~\ref{tbl:movielens_genre_1} \&~\ref{tbl:movielens_genre_2} and Amazon Review product shifts in Table~\ref{tbl:amazon_product}. We compare \levi{} with 1) standard training baselines (e.g., full fine-tuning, light-tuning, and training-from-scratch) and 2) state-of-the-art fine-tuning generalization baselines.

The results are consistent with those in Table~\ref{tbl:recom_main}. For example, in many cases, full fine-tuned models show worse OOD results than either the light-tuned or trained-from-scratch models. This phenomenon is especially notable in Table~\ref{tbl:amazon_product}, where the fine-tuned model performs far worse than all the light-tuned and trained-from-scratch models.
Also, the fine-tuning generalization baselines (e.g., LP$\rightarrow$FT, FT+RobustModel) mostly improve the OOD performances of the fine-tuned model. However, the FT+ZeroShot baseline sometimes largely fails to improve the OOD performances, as in Table~\ref{tbl:amazon_product}. As we discussed in Sec.~\ref{sec:performances_id_ood}, we suspect that the pre-trained (zero-shot) language features are not sufficient for this downstream task and thus may even harm the results.
In comparison, \levi{} further improves OOD results in all types of shifts. 

\begin{table}[h]
  \caption{OOD performances on the MovieLens dataset, where the type of OOD is genre shifts (part 1).
% \vspace{-0.2cm}
  }
  \label{tbl:movielens_genre_1}
  \centering
\scalebox{0.9}{
  \begin{tabular}{lcccccc}
    \toprule
      Method & Adventure & Sci-Fi & War & Crime & Children & Horror \\
    \cmidrule(lr){1-1}\cmidrule(lr){2-7}
    T5x Fine-tuning (FT) & 1.251\tiny{$\pm$0.036} & 1.314\tiny{$\pm$0.046} & 1.153\tiny{$\pm$0.033} & 1.341\tiny{$\pm$0.005} & 1.262\tiny{$\pm$0.045} & 1.355\tiny{$\pm$0.050} \\
    T5x Light-tuning: Half of transformers (HT) & 1.283\tiny{$\pm$0.018} & 1.352\tiny{$\pm$0.022} & 1.276\tiny{$\pm$0.019} & 1.405\tiny{$\pm$0.089} & 1.298\tiny{$\pm$0.028} & 1.403\tiny{$\pm$0.008} \\
    T5x Light-tuning: Linear probing (LP) & 1.233\tiny{$\pm$0.005} & 1.231\tiny{$\pm$0.099} & 1.166\tiny{$\pm$0.224} & 1.402\tiny{$\pm$0.118} & 1.195\tiny{$\pm$0.032} & 1.398\tiny{$\pm$0.123} \\
    T5x From scratch (FS) & 1.350\tiny{$\pm$0.022} & 1.296\tiny{$\pm$0.018} & 1.246\tiny{$\pm$0.032} & 1.367\tiny{$\pm$0.060} & 1.279\tiny{$\pm$0.031} & 1.399\tiny{$\pm$0.046} \\
    \cmidrule(lr){1-1}\cmidrule(lr){2-7}
    LP$\rightarrow$FT~\citep{kumar2022finetuning} & 1.335\tiny{$\pm$0.008} & 1.352\tiny{$\pm$0.013} & 1.294\tiny{$\pm$0.071} & 1.340\tiny{$\pm$0.002} & 1.368\tiny{$\pm$0.030} & 1.378\tiny{$\pm$0.004} \\
    FT+RobustModel~\citep{kumar2022calibrated} & 1.122\tiny{$\pm$0.013} & 1.175\tiny{$\pm$0.045} & 1.066\tiny{$\pm$0.068} & 1.203\tiny{$\pm$0.026} & 1.103\tiny{$\pm$0.013} & 1.257\tiny{$\pm$0.063} \\
    FT+FS & 1.156\tiny{$\pm$0.018} & 1.196\tiny{$\pm$0.024} & 1.076\tiny{$\pm$0.002} & 1.235\tiny{$\pm$0.010} & 1.142\tiny{$\pm$0.032} & 1.259\tiny{$\pm$0.042} \\
    FT+ZeroShot~\citep{wortsman2022robust} & 1.227\tiny{$\pm$0.032} & 1.338\tiny{$\pm$0.036} & 1.134\tiny{$\pm$0.020} & 1.343\tiny{$\pm$0.003} & 1.239\tiny{$\pm$0.038} & 1.363\tiny{$\pm$0.029} \\
    \cmidrule(lr){1-1}\cmidrule(lr){2-7}
    \levi{} & \textbf{1.036\tiny{$\pm$0.006}} & \textbf{1.138\tiny{$\pm$0.010}} & \textbf{1.059\tiny{$\pm$0.069}} & \textbf{1.178\tiny{$\pm$0.051}} & \textbf{1.047\tiny{$\pm$0.017}} & \textbf{1.105\tiny{$\pm$0.019}} \\
    % \levi{} with original T5x & 1.074\tiny{$\pm$0.024} & 1.180\tiny{$\pm$0.025} & 1.260\tiny{$\pm$0.104} & \textbf{1.097\tiny{$\pm$0.087}} & 1.041\tiny{$\pm$0.021} & 1.146\tiny{$\pm$0.023} \\
    % \levi{} with light-tuned T5x & 1.036\tiny{$\pm$0.006} & \textbf{1.138\tiny{$\pm$0.010}} & \textbf{1.059\tiny{$\pm$0.069}} & 1.178\tiny{$\pm$0.051} & 1.047\tiny{$\pm$0.017} & \textbf{1.105\tiny{$\pm$0.019}} \\
    % \levi{} with full fine-tuned T5x & \textbf{1.016\tiny{$\pm$0.013}} & 1.164\tiny{$\pm$0.011} & 1.097\tiny{$\pm$0.015} & 1.128\tiny{$\pm$0.047} & \textbf{1.040\tiny{$\pm$0.006}} & 1.129\tiny{$\pm$0.016} \\
    \bottomrule
  \end{tabular}
  }
%   \vspace{-0.1cm}
\end{table}

\begin{table}[h]
  \caption{OOD performances on the MovieLens dataset, where the type of OOD is genre shifts (part 2).
% \vspace{-0.2cm}
  }
  \label{tbl:movielens_genre_2}
  \centering
\scalebox{0.9}{
  \begin{tabular}{lcccccc}
    \toprule
      Method &  Mystery & Musical & Animation & Western & Film-Noir & Documentary \\
    \cmidrule(lr){1-1}\cmidrule(lr){2-7}
    T5x Fine-tuning (FT) & 1.268\tiny{$\pm$0.023} & 1.265\tiny{$\pm$0.060} & 1.262\tiny{$\pm$0.049} & 1.216\tiny{$\pm$0.068} & 1.125\tiny{$\pm$0.082} & 1.404\tiny{$\pm$0.045} \\
    T5x Light-tuning: Half of transformers (HT) & 1.245\tiny{$\pm$0.038} & 1.315\tiny{$\pm$0.024} & 1.327\tiny{$\pm$0.023} & 1.194\tiny{$\pm$0.013} & 1.122\tiny{$\pm$0.014} & 1.467\tiny{$\pm$0.031} \\
    T5x Light-tuning: Linear probing (LP) & 1.213\tiny{$\pm$0.122} & 1.158\tiny{$\pm$0.025} & 1.166\tiny{$\pm$0.015} & 1.100\tiny{$\pm$0.101} & 1.062\tiny{$\pm$0.273} & 1.288\tiny{$\pm$0.062} \\
    T5x From scratch (FS)  & 1.303\tiny{$\pm$0.043} & 1.259\tiny{$\pm$0.031} & 1.262\tiny{$\pm$0.031} & 1.187\tiny{$\pm$0.038} & 1.160\tiny{$\pm$0.025} & 1.388\tiny{$\pm$0.044} \\
    \cmidrule(lr){1-1}\cmidrule(lr){2-7}
    LP$\rightarrow$FT~\citep{kumar2022finetuning} & 1.428\tiny{$\pm$0.083} & 1.356\tiny{$\pm$0.049} & 1.396\tiny{$\pm$0.034} & 1.327\tiny{$\pm$0.039} & 1.260\tiny{$\pm$0.004} & 1.455\tiny{$\pm$0.029} \\
    FT+RobustModel~\citep{kumar2022calibrated} & 1.125\tiny{$\pm$0.084} & 1.097\tiny{$\pm$0.028} & 1.092\tiny{$\pm$0.020} & 1.048\tiny{$\pm$0.088} & 1.017\tiny{$\pm$0.127} & \textbf{1.221\tiny{$\pm$0.073}} \\
    FT+FS & 1.167\tiny{$\pm$0.033} & 1.143\tiny{$\pm$0.028} & 1.139\tiny{$\pm$0.026} & 1.091\tiny{$\pm$0.044} & 1.033\tiny{$\pm$0.018} & 1.277\tiny{$\pm$0.024} \\
    FT+ZeroShot~\citep{wortsman2022robust} & 1.254\tiny{$\pm$0.016} & 1.235\tiny{$\pm$0.049} & 1.246\tiny{$\pm$0.033} & 1.124\tiny{$\pm$0.096} & 1.143\tiny{$\pm$0.076} & 1.396\tiny{$\pm$0.026} \\
    \cmidrule(lr){1-1}\cmidrule(lr){2-7}
    \levi{} & \textbf{0.973\tiny{$\pm$0.013}} & \textbf{1.077\tiny{$\pm$0.031}} & \textbf{1.074\tiny{$\pm$0.015}} & \textbf{0.949\tiny{$\pm$0.029}} & \textbf{0.878\tiny{$\pm$0.052}} & 1.267\tiny{$\pm$0.031} \\
    % \levi{} with original T5x & 1.072\tiny{$\pm$0.056} & \textbf{1.052\tiny{$\pm$0.042}} & \textbf{1.065\tiny{$\pm$0.007}} & 1.056\tiny{$\pm$0.100} & 0.959\tiny{$\pm$0.034} & 1.251\tiny{$\pm$0.072} \\
    % \levi{} with light-tuned T5x & \textbf{0.973\tiny{$\pm$0.013}} & 1.077\tiny{$\pm$0.031} & 1.074\tiny{$\pm$0.015} & \textbf{0.949\tiny{$\pm$0.029}} & \textbf{0.878\tiny{$\pm$0.052}} & 1.267\tiny{$\pm$0.031} \\
    % \levi{} with full fine-tuned T5x & 1.051\tiny{$\pm$0.017} & 1.067\tiny{$\pm$0.024} & 1.074\tiny{$\pm$0.015} & 1.044\tiny{$\pm$0.058} & 0.975\tiny{$\pm$0.037} & 1.240\tiny{$\pm$0.042} \\
    \bottomrule
  \end{tabular}
  }
\vspace{0.2cm}
\end{table}

\begin{table}[h]
\vspace{0.3cm}
  \caption{OOD performances on the Amazon Review dataset, where the type of OOD is product shifts.
% \vspace{-0.2cm}
  }
  \label{tbl:amazon_product}
  \centering
\scalebox{0.9}{
% \hspace{-0.3cm}
  \begin{tabular}{l@{\hspace{12pt}}c@{\hspace{9pt}}c@{\hspace{9pt}}c@{\hspace{9pt}}c@{\hspace{9pt}}c@{\hspace{9pt}}c@{\hspace{9pt}}c}
    \toprule
    %   & & \multicolumn{12}{c}{OOD} \\
    % \cmidrule(lr){1-1}\cmidrule(lr){2-3}\cmidrule(lr){4-5}
      Method & Watch & Toy & Sports & Music & Jewelry & Furniture & Baby \\
    \cmidrule(lr){1-1}\cmidrule(lr){2-8}
    T5x Fine-tuning (FT) & 2.016\tiny{$\pm$0.131} & 2.015\tiny{$\pm$0.181} & 2.024\tiny{$\pm$0.165} & 1.986\tiny{$\pm$0.164} & 2.029\tiny{$\pm$0.158} & 2.075\tiny{$\pm$0.125} & 2.057\tiny{$\pm$0.151} \\
    % \cmidrule(lr){1-1}\cmidrule(lr){2-8}
    T5x Light-tuning: Half of transformers (HT) & 1.925\tiny{$\pm$0.291} & 1.875\tiny{$\pm$0.304} & 1.814\tiny{$\pm$0.287} & 1.645\tiny{$\pm$0.295} & 1.966\tiny{$\pm$0.239} & 1.916\tiny{$\pm$0.304} & 1.892\tiny{$\pm$0.303} \\
    T5x Light-tuning: Linear probing (LP) & 1.776\tiny{$\pm$0.120} & 1.782\tiny{$\pm$0.129} & 1.752\tiny{$\pm$0.135} & 1.668\tiny{$\pm$0.165} & 1.763\tiny{$\pm$0.123} & 1.792\tiny{$\pm$0.110} & 1.782\tiny{$\pm$0.125} \\
    T5x From scratch (FS) & 1.564\tiny{$\pm$0.004} & 1.505\tiny{$\pm$0.007} & 1.471\tiny{$\pm$0.012} & 1.241\tiny{$\pm$0.029} & 1.553\tiny{$\pm$0.004} & 1.593\tiny{$\pm$0.003} & 1.532\tiny{$\pm$0.012} \\
    \cmidrule(lr){1-1}\cmidrule(lr){2-8}
    LP$\rightarrow$FT~\citep{kumar2022finetuning} & 1.692\tiny{$\pm$0.023} & 1.693\tiny{$\pm$0.032} & 1.657\tiny{$\pm$0.026} & 1.551\tiny{$\pm$0.010} & 1.676\tiny{$\pm$0.022} & 1.716\tiny{$\pm$0.034} & 1.696\tiny{$\pm$0.029} \\
    FT+RobustModel~\citep{kumar2022calibrated} & 1.757\tiny{$\pm$0.111} & 1.754\tiny{$\pm$0.141} & 1.745\tiny{$\pm$0.139} & 1.686\tiny{$\pm$0.162} & 1.759\tiny{$\pm$0.123} & 1.789\tiny{$\pm$0.108} & 1.776\tiny{$\pm$0.129} \\
    FT+FS & 1.561\tiny{$\pm$0.015} & 1.526\tiny{$\pm$0.031} & 1.506\tiny{$\pm$0.031} & 1.341\tiny{$\pm$0.047} & 1.562\tiny{$\pm$0.024} & 1.604\tiny{$\pm$0.022} & 1.550\tiny{$\pm$0.031} \\
    FT+ZeroShot~\citep{wortsman2022robust} & 2.509\tiny{$\pm$0.149} & 2.467\tiny{$\pm$0.169} & 2.527\tiny{$\pm$0.144} & 2.647\tiny{$\pm$0.150} & 2.481\tiny{$\pm$0.176} & 2.478\tiny{$\pm$0.218} & 2.498\tiny{$\pm$0.221} \\
    \cmidrule(lr){1-1}\cmidrule(lr){2-8}
    \levi{} & \textbf{1.346\tiny{$\pm$0.003}} & \textbf{1.303\tiny{$\pm$0.003}} & \textbf{1.275\tiny{$\pm$0.003}} & \textbf{1.118\tiny{$\pm$0.013}} & \textbf{1.348\tiny{$\pm$0.007}} & \textbf{1.366\tiny{$\pm$0.010}} & \textbf{1.327\tiny{$\pm$0.002}} \\
    % \midrule
    % \levi{} with original T5x & \textbf{1.332\tiny{$\pm$0.011}} & \textbf{1.289\tiny{$\pm$0.003}} & \textbf{1.254\tiny{$\pm$0.005}} & \textbf{1.068\tiny{$\pm$0.003}} & \textbf{1.327\tiny{$\pm$0.009}} & \textbf{1.350\tiny{$\pm$0.008}} & \textbf{1.303\tiny{$\pm$0.005}} \\
    % \levi{} with light-tuned T5x & \textbf{1.346\tiny{$\pm$0.003}} & \textbf{1.303\tiny{$\pm$0.003}} & \textbf{1.275\tiny{$\pm$0.003}} & \textbf{1.118\tiny{$\pm$0.013}} & \textbf{1.348\tiny{$\pm$0.007}} & \textbf{1.366\tiny{$\pm$0.010}} & \textbf{1.327\tiny{$\pm$0.002}} \\
    % \levi{} with full fine-tuned T5x & 1.539\tiny{$\pm$0.101} & 1.513\tiny{$\pm$0.105} & 1.554\tiny{$\pm$0.114} & 1.443\tiny{$\pm$0.149} & 1.522\tiny{$\pm$0.084} & 1.542\tiny{$\pm$0.098} & 1.556\tiny{$\pm$0.121} \\
    \bottomrule
  \end{tabular}
  }
\vspace{0.3cm}
\end{table}

\vspace{0.5cm}
\subsection{More Results on Vision Tasks}
\label{appx:vision_all}

Continuing from Sec.~\ref{sec:performances_id_ood}, we provide full OOD performances on the ImageNet-variant datasets in Table~\ref{tbl:imagenet}. We use the same baselines in the previous section except the trained-from-scratch baseline, as ViT models are known to be hard to train on small or mid-sized training data using random weight initialization~\citep{steiner2022how}.

As we observed in Sec.~\ref{sec:performances_id_ood}, the existing fine-tuning generalization baselines improve OOD performances, as the ImageNet pre-trained ViT has reasonable features to support ImageNet-variants datasets. Although these baselines show more effective results in the ImageNet scenario compared to the language-based recommendation tasks, \levi{} still achieves the best or second-best OOD accuracies among all baselines while having ID accuracies comparable to that of the fine-tuned model -- see the ID results in Table~\ref{tbl:vision_main}.

\begin{table}[h]
  \caption{OOD performances on the ImageNet-variant datasets: ImageNet-A, ImageNet-R, and ImageNet-V2.
% \vspace{-0.2cm}
  }
  \label{tbl:imagenet}
  \centering
\scalebox{0.95}{
% \hspace{-0.3cm}
  \begin{tabular}{lccc}
    \toprule
      Method &  OOD Acc: A & OOD Acc: R & OOD Acc: V2 \\
    \cmidrule(lr){1-1}\cmidrule(lr){2-4}
    ViT Fine-tuning (FT) & 1.90\tiny{$\pm$0.22} & 33.60\tiny{$\pm$0.74} & 29.37\tiny{$\pm$0.49}\\
    % \cmidrule(lr){1-1}\cmidrule(lr){2-4}
    ViT Light-tuning: Half of transformers (HT) & 4.46\tiny{$\pm$0.26} & 38.48\tiny{$\pm$0.21} & 43.62\tiny{$\pm$0.45} \\
    ViT Light-tuning: Linear probing (LP) & \textbf{8.57\tiny{$\pm$0.04}} & 35.59\tiny{$\pm$0.05} & \textbf{50.32\tiny{$\pm$0.12}}\\
    \cmidrule(lr){1-1}\cmidrule(lr){2-4}
    LP$\rightarrow$FT~\citep{kumar2022finetuning} & 4.43\tiny{$\pm$0.19} & 36.90\tiny{$\pm$0.15} & 43.44\tiny{$\pm$0.15}\\
    FT+RobustModel~\citep{kumar2022calibrated} & 5.22\tiny{$\pm$0.08} & 39.98\tiny{$\pm$0.54} & 45.71\tiny{$\pm$0.15}\\
    FT+ZeroShot~\citep{wortsman2022robust} & 5.21\tiny{$\pm$0.03} & \underline{43.35\tiny{$\pm$0.16}} & 47.75\tiny{$\pm$0.32}\\
    \cmidrule(lr){1-1}\cmidrule(lr){2-4}
    \levi{} & \underline{5.25\tiny{$\pm$0.02}} & \textbf{45.94\tiny{$\pm$0.19}} & \underline{49.68\tiny{$\pm$0.14}}\\
    % \levi{} with original ViT & x.xx\tiny{$\pm$x.xx} & xx.xx\tiny{$\pm$x.xx} & xx.xx\tiny{$\pm$x.xx}\\
    % \levi{} with light-tuned ViT & \underline{5.82\tiny{$\pm$0.01}} & 41.74\tiny{$\pm$0.18} & 46.64\tiny{$\pm$0.02}\\
    % \levi{} with full fine-tuned ViT & 5.25\tiny{$\pm$0.02} & \textbf{45.94\tiny{$\pm$0.19}} & \underline{49.68\tiny{$\pm$0.14}}\\
    % \levi{} with full fine-tuned ViT & 4.73\tiny{$\pm$0.34} & 45.48\tiny{$\pm$0.33} & 49.05\tiny{$\pm$0.41}\\
    % \levi{} with full fine-tuned ViT & 5.23\tiny{$\pm$0.04} & 46.61\tiny{$\pm$0.03} & 50.07\tiny{$\pm$0.20}\\
    \bottomrule
  \end{tabular}
  }
  \vspace{0.5cm}
\end{table}

\newpage
\subsection{\yuji{Results on Another NLP Task}}
\label{appx:nlp}

\yuji{Continuing from Sec.~\ref{sec:experiments}, we perform an additional experiment for sentiment classification, a different type of natural language processing (NLP) tasks, and observe consistent performance improvements when using LEVI.}

\yuji{In this experiment, we revisit the Amazon Review dataset~\citep{ni2019justifying}, which can be used for the sentiment analysis of customers using their review texts~\citep{xie2020unsupervised,du2019explicit}. We note that while ``review texts'' have not been used as an input feature in our recommendation experiments to follow the common setting of the recommendation task, we now use the text information as the main input feature for classifying customer sentiments. For the baselines, we compare four standard training baselines (FT, HT, LP, and FS) and a state-of-the-art fine-tuning generalization baseline FT+RobustModel~\citep{kumar2022calibrated}, which shows the best performances among the baselines in Table~\ref{tbl:recom_main}.
}

\yuji{
As a result, Table~\ref{tbl:nlp} shows that LEVI outperforms all the baselines in terms of both ID and OOD performances in the sentiment classification task. This result indicates that LEVI can support more general types of NLP tasks.
}

\begin{table}[h]
  \caption{\yuji{Performances on the Amazon Review dataset for sentiment classification.}
  \vspace{0.1cm}
  }
  \label{tbl:nlp}
  \centering
\scalebox{0.95}{
% \hspace{-0.3cm}
  \begin{tabular}{lcc}
    \toprule
      Method &  ID RMSE & OOD RMSE \\
    \cmidrule(lr){1-1}\cmidrule(lr){2-3}
    T5x Fine-tuning (FT) & 0.686 & 0.802\\
    T5x Light-tuning: Half of transformers (HT) & 0.743 & 0.868\\
    T5x Light-tuning: Linear probing (LP) & 1.343 & 1.487\\
    T5x From scratch (FS) & 0.900 & 1.089\\
    FT+RobustModel~\citep{kumar2022calibrated} & 0.624 & 0.727\\
    \cmidrule(lr){1-1}\cmidrule(lr){2-3}
    \levi{} & \textbf{0.587} & \textbf{0.691}\\
    \bottomrule
  \end{tabular}
  }
  \vspace{0.5cm}
\end{table}

% \newpage
\vspace{0.5cm}
\subsection{\levi{} with Different Large Models}
\label{appx:different_largemodels}

Continuing from Secs.~\ref{sec:performances_id_ood} and~\ref{appx:model_settings}, we provide the results of \levi{} with pre-trained, light-tuned, and fine-tuned models. As \levi{}'s large model can be one of original pre-trained or downstream task adapted (i.e., light-tuned or fine-tuned) models, we compare the performances of \levi{} with different large models.

Table~\ref{tbl:levi_different_models} shows the ID and OOD performances of \levi{} on the MovieLens and Amazon Review datasets. The ID performances are better when using the fully fine-tuned large models in \levi{} compared to using the pre-trained and light-tuned models. In comparison, the OOD performances are better when using the pre-trained and light-tuned models. This result shows the relationship between the ID-OOD tradeoff and adapting large models to the fine-tuning (ID) data. \levi{} can enjoy better ID performances when using a large model more adapted to the fine-tuning (ID) data, while achieving better OOD performances with an original pre-trained model, which is less affected by the problematic features in the fine-tuning data.

\begin{table*}[h]
  \caption{Performances of \levi{} on the MovieLens and Amazon Review datasets with different large models.
  }
  \label{tbl:levi_different_models}
  \centering
\scalebox{0.95}{
% \hspace{-0.3cm}
  \begin{tabular}{l@{\hspace{12pt}}c@{\hspace{9pt}}c@{\hspace{12pt}}c@{\hspace{9pt}}c@{\hspace{9pt}}c}
    \toprule
      & \multicolumn{2}{c}{MovieLens} & \multicolumn{3}{c}{Amazon Review} \\
    \cmidrule(lr){1-1}\cmidrule(lr){2-3}\cmidrule(lr){4-6}
      Method &  ID & OOD (12 genres) & ID & OOD (time) & OOD (7 products) \\
    \cmidrule(lr){1-1}\cmidrule(lr){2-3}\cmidrule(lr){4-6} 
    \levi{} with pre-trained T5x & 0.944\tiny{$\pm$0.004} & 1.104\tiny{$\pm$0.003} & 1.109\tiny{$\pm$0.003} & \textbf{1.296\tiny{$\pm$0.011}} & \textbf{1.275\tiny{$\pm$0.006}}\\
    \levi{} with light-tuned T5x & 0.932\tiny{$\pm$0.005} & \textbf{1.065\tiny{$\pm$0.018}} & 1.095\tiny{$\pm$0.003} & 1.310\tiny{$\pm$0.006} & 1.298\tiny{$\pm$0.006}\\
    \levi{} with fully fine-tuned T5x & \textbf{0.927\tiny{$\pm$0.004}} & 1.085\tiny{$\pm$0.014} & \textbf{1.093\tiny{$\pm$0.001}} & 1.318\tiny{$\pm$0.004} & 1.524\tiny{$\pm$0.110}\\
    \bottomrule
  \end{tabular}
  }
%   \vspace{-0.4cm}
\end{table*}

% \vspace{0.5cm}
\newpage
\subsection{More Results on Efficiency Comparison}
\label{appx:efficiency_all}

Continuing from Sec.~\ref{sec:efficiency}, we provide the full results of efficiency comparison between \levi{} and baselines. 
We use 1) the number of model parameters and 2) floating point operations (FLOPs).
Tables~\ref{tbl:efficiency_t5x} and~\ref{tbl:efficiency_vit} show the comparisons for T5x and ViT, respectively.
As we observed in Sec.~\ref{sec:efficiency}, when \levi{} uses a pre-trained model, the number of training parameters is much lower than all state-of-the-art baselines, and the other two metrics are comparable to those of using a single large model. When \levi{} uses a fine-tuned model, it shows comparable results with single model-based baselines, including FT+ZeroShot, in all three metrics. Compared to the heavy ensembles with two large models (i.e., FT+RobustModel, FT+FS), which show good performances among the baselines in various tasks, \levi{} is much efficient while also achieving better OOD generalization.
% Compared to the heavy ensembles (i.e., FT+RobustModel, FT+FS), which consistently show good performances among the baselines, \levi{} performs training and inference much faster while also achieving better OOD generalization.

\begin{table}[h]
\vspace{0.3cm}
  \caption{Number of parameters and FLOPs of baselines and \levi{} with T5x. The FLOPs of the default T5x model are obtained from the previous study~\citep{akbari-etal-2022-e}.
  }
  \vspace{0.1cm}
  \label{tbl:efficiency_t5x}
  \centering
  \scalebox{0.95}{
  \begin{tabular}{lccc}
    \toprule
    %   &  \multicolumn{3}{c}{T5x} &\multicolumn{3}{c}{ViT} \\
    Method & Params (Train.) & Params (Infer.) & FLOPs\\
    \cmidrule(lr){1-1}\cmidrule(lr){2-4}
    Fine-tuning (FT) & 60M & 60M & 33G \\
    Light-tuning: Half of transformers (HT) & 30M & 60M & 33G \\
    Light-tuning: Linear probing (LP) & $\sim$5K & 60M & 33G \\
    From scratch (FS) & 60M & 60M & 33G \\
    \cmidrule(lr){1-1}\cmidrule(lr){2-4}
    LP$\rightarrow$FT~\citep{kumar2022finetuning} & 60M + $\sim$5K & 60M & 33G \\
    FT+RobustModel~\citep{kumar2022calibrated} & 120M & 120M & 66G \\
    FT+FS & 120M & 120M & 66G \\
    FT+ZeroShot~\citep{wortsman2022robust} & 60M & 60M & 33G \\
    \cmidrule(lr){1-1}\cmidrule(lr){2-4}
    \levi{} with pre-trained model & $\sim$2M & 60M + $\sim$2M & 33G + $\sim$4M \\
    \levi{} with light-tuned model & 30M + $\sim$2M & 60M + $\sim$2M & 33G + $\sim$4M \\
    \levi{} with fully fine-tuned model & 60M + $\sim$2M & 60M + $\sim$2M & 33G + $\sim$4M \\
    \bottomrule
  \end{tabular}
  }
  \vspace{0.5cm}
\end{table}

\begin{table}[h]
  \caption{Number of parameters and FLOPs of baselines and \levi{} with ViT. The FLOPs of the default ViT model are obtained from the previous study~\citep{rao2021dynamicvit}.
  }
  \vspace{0.1cm}
  \label{tbl:efficiency_vit}
  \centering
  \scalebox{0.95}{
  \begin{tabular}{lccc}
    \toprule
    %   &  \multicolumn{3}{c}{T5x} &\multicolumn{3}{c}{ViT} \\
    Method & Params (Train.) & Params (Infer.) & FLOPs \\
    \cmidrule(lr){1-1}\cmidrule(lr){2-4}
    Fine-tuning (FT) & 86M & 86M & 17.5G \\
    Light-tuning: Half of transformers (HT) & 43M & 86M & 17.5G \\
    Light-tuning: Linear probing (LP) & $\sim$1M & 86M & 17.5G \\
    % From scratch (FS)  & - & - & - \\
    \cmidrule(lr){1-1}\cmidrule(lr){2-4}
    LP$\rightarrow$FT~\citep{kumar2022finetuning} & 86M + $\sim$1M & 86M & 17.5G\\
    FT+RobustModel~\citep{kumar2022calibrated} & 172M & 172M & 35G\\
    % FT+FS & - & - & - \\
    FT+ZeroShot~\citep{wortsman2022robust} & 86M & 86M & 17.5G\\
    \cmidrule(lr){1-1}\cmidrule(lr){2-4}
    \levi{} with pre-trained model & $\sim$15M & 86M + $\sim$15M & 17.5G + $\sim$5G\\
    \levi{} with light-tuned model & 43M + $\sim$15M & 86M + $\sim$15M & 17.5G + $\sim$5G\\
    \levi{} with fully fine-tuned model & 86M + $\sim$15M & 86M + $\sim$15M & 17.5G + $\sim$5G\\
    \bottomrule
  \end{tabular}
  }
\vspace{0.5cm}
\end{table}

\newpage
% \vspace{0.5cm}
\subsection{More Results on Compatibility with Efficient Fine-Tuning Methods}
\label{appx:lora_all}

Continuing from Sec.~\ref{sec:compatibility}, we provide full OOD performances when using LoRA~\citep{hu2022lora} on the ImageNet-variant datasets. When \levi{} uses a fine-tuned model instead of an original pre-trained model, we can use efficient fine-tuning techniques like LoRA to improve the overall training efficiency.
Table~\ref{tbl:lora_imagenet} shows that \levi{} improves the OOD performances of the LoRA-tuned models in all the three ImageNet-variant datasets (ImageNet-A, ImageNet-R, and ImageNet-V2), demonstrating that \levi{} can be gracefully merged together with existing efficient fine-tuning methods.

\begin{table}[h]
\vspace{0.3cm}
  \caption{OOD performances of LoRA experiments on the ImageNet-variant datasets: ImageNet-A, ImageNet-R, and ImageNet-V2.
\vspace{0.1cm}
  }
  \label{tbl:lora_imagenet}
  \centering
\scalebox{0.95}{
% \hspace{-0.3cm}
  \begin{tabular}{lccc}
    \toprule
      Method &  OOD Acc: A & OOD Acc: R & OOD Acc: V2 \\
    \cmidrule(lr){1-1}\cmidrule(lr){2-4}
    LoRA-tuned ViT & 2.50\tiny{$\pm$0.01} & 31.21\tiny{$\pm$0.36} & 32.53\tiny{$\pm$0.14}\\
    \levi{} with LoRA-tuned ViT & \textbf{3.76\tiny{$\pm$0.09}} & \textbf{35.91\tiny{$\pm$0.38}} & \textbf{37.25\tiny{$\pm$0.27}}\\
    \bottomrule
  \end{tabular}
  }
%   \vspace{-0.1cm}
\end{table}

\vspace{0.5cm}
\subsection{More Results on Compatibility with Efficient Fine-Tuning Methods}
\label{appx:lora_all}

Continuing from Sec.~\ref{sec:compatibility}, we provide full OOD performances when using LoRA~\citep{hu2022lora} on the ImageNet-variant datasets. When \levi{} uses a fine-tuned model instead of an original pre-trained model, we can use efficient fine-tuning techniques like LoRA to improve the overall training efficiency.
Table~\ref{tbl:lora_imagenet} shows that \levi{} improves the OOD performances of the LoRA-tuned models in all the three ImageNet-variant datasets (ImageNet-A, ImageNet-R, and ImageNet-V2), demonstrating that \levi{} can be gracefully merged together with existing efficient fine-tuning methods.

\begin{table}[h]
\vspace{0.3cm}
  \caption{OOD performances of LoRA experiments on the ImageNet-variant datasets: ImageNet-A, ImageNet-R, and ImageNet-V2.
\vspace{0.1cm}
  }
  \label{tbl:lora_imagenet}
  \centering
\scalebox{0.95}{
% \hspace{-0.3cm}
  \begin{tabular}{lccc}
    \toprule
      Method &  OOD Acc: A & OOD Acc: R & OOD Acc: V2 \\
    \cmidrule(lr){1-1}\cmidrule(lr){2-4}
    LoRA-tuned ViT & 2.50\tiny{$\pm$0.01} & 31.21\tiny{$\pm$0.36} & 32.53\tiny{$\pm$0.14}\\
    \levi{} with LoRA-tuned ViT & \textbf{3.76\tiny{$\pm$0.09}} & \textbf{35.91\tiny{$\pm$0.38}} & \textbf{37.25\tiny{$\pm$0.27}}\\
    \bottomrule
  \end{tabular}
  }
%   \vspace{-0.1cm}
\end{table}

\vspace{0.5cm}
\subsection{\yuji{Illustration on Spurious Feature Mitigation}}
\label{appx:evidence}

\yuji{Continuing from Sec.~\ref{sec:experiments}, we demonstrate that \levi{} can mitigate the impact of spurious features. 
For a clear illustration, we perform a synthetic experiment with datasets consisting of two spurious features (s1, s2), one transferable feature (x), and one binary label attribute (y). Specifically, we construct three datasets: 1) a pre-training dataset, where s1 and x are correlated with y; 2) a fine-tuning dataset, where s2 and x are correlated with y; and 3) a test dataset, where only x is correlated with y. All datasets have 2,000 data points.}

\yuji{We compare three models, 1) fine-tuned, 2) trained-from-scratch, and 3) LEVI-based models. 1) We first pre-train a 3-layer neural network on the given pre-training data and then fine-tune it on the fine-tuning data. 2) We also prepare a trained-from-scratch model only using the fine-tuning data. 3) Finally, we train a LEVI model that uses complementing views from the pre-trained model and the trained-from-scratch model.}

\yuji{When applying the three models on the test data (only has the transferable feature), we have the following result: the fine-tuned, trained-from-scratch, and LEVI models achieve accuracies (the higher the better) of 68.9, 69.5, and 75.5, respectively. This result indicates that LEVI is clearly less affected by the spurious correlations by s1 and s2 and uses the information of transferable feature x, compared to the fine-tuned and trained-from-scratch models.}

\newpage
\section{Possible Extension via Other Efficient Ensemble Techniques}
\label{appx:discussion}
Continuing from Sec.~\ref{sec:compatibility}, we provide more discussion on a possible extension of \levi{} to further improve its efficiency via other efficient ensemble techniques.
Here, we focus on a widely used efficient ensemble technique called BatchEnsemble~\citep{wen2020batchensemble}, which is designed to reduce the computational and memory costs of typical heavy ensembles of multiple models, while achieving similar performances with the heavy ensembles. The key idea of BatchEnsemble is to use one shared weight matrix with multiple rank-one matrices, where each rank-one matrix is multiplied by the shared matrix to recover each member (model) of the original ensemble -- please refer to the details in \citet{wen2020batchensemble}. 

To further improve the efficiency of \levi{}, we can consider applying the idea of BatchEnsemble, especially to the adapting layers in \levi{}. When downstream tasks and models become much more complex and large, \levi{} may require larger-sized adapting layers compared to those used in our current experiments. In such cases, one possible way to reduce the computational and memory costs can be to set each adapting layer with a single-rank matrix, as in BatchEnsemble. Although \levi{} is already very efficient compared to existing heavy ensembles, extending our work by studying the compatibility with other efficient ensemble methods will be a promising future direction.

\end{document}